\pdfoutput=1
\documentclass[11pt]{article}

\usepackage[final]{acl}

\usepackage{times}
\usepackage{latexsym}

\usepackage[T1]{fontenc}
\usepackage[utf8]{inputenc}
\usepackage{microtype}
\usepackage{inconsolata}
\usepackage{fontawesome}
\usepackage{pgfplots}
\pgfplotsset{compat=1.18}
\usepackage{xcolor}
\usepackage{graphicx}
\graphicspath{{figs/}}

\usepackage{amsmath}
\usepackage{amssymb}
\usepackage{algorithm}
\usepackage{algpseudocode}
\usepackage{placeins}
\usepackage[most]{tcolorbox}
\usepackage{booktabs}
\usepackage{tabularx}
\usepackage{makecell}
\usepackage{multirow}
\usepackage{xcolor}
\usepackage{colortbl}
\usepackage{adjustbox}
\usepackage{caption}
\usepackage{siunitx}
\usepackage{pifont}
\usepackage{needspace}
\usepackage{pgfplots}
\definecolor{barNeutralLt}{HTML}{E9EAED}
\definecolor{barNeutral}{HTML}{AEB3BC}
\definecolor{barBlueMid}{HTML}{6E91B8}
\definecolor{barGreenLt}{HTML}{BFD9BC}
\definecolor{barGreen}{HTML}{7CB877}
\definecolor{barGreenDk}{HTML}{3C7A38}
\pgfplotsset{compat=1.18}
\usetikzlibrary{positioning,arrows.meta,fit,patterns}

\definecolor{deltaup}{RGB}{0,135,40}
\definecolor{deltadn}{RGB}{200,30,30}
\definecolor{amberhi}{RGB}{220,140,30}
\definecolor{amberlt}{RGB}{255,232,200}
\definecolor{bandgray}{RGB}{235,235,238}
\definecolor{bandblue}{RGB}{225,235,250}
\definecolor{bandred}{RGB}{250,228,228}
\definecolor{bandgreen}{RGB}{225,245,225}
\definecolor{barblue}{RGB}{90,135,200}
\definecolor{barblueLt}{RGB}{180,205,240}
\definecolor{barneg}{RGB}{240,170,170}
\definecolor{barNeutral}{RGB}{160,165,175}
\definecolor{barNeutralLt}{RGB}{215,218,225}
\definecolor{barBlueMid}{RGB}{80,125,190}
\definecolor{barGreen}{RGB}{60,150,90}
\definecolor{barGreenLt}{RGB}{160,210,170}
\definecolor{barGreenDk}{RGB}{40,120,70}
\definecolor{barRed}{RGB}{195,80,80}

\newcommand{\dn}[1]{\textsubscript{\textcolor{deltadn}{$\downarrow$\,#1}}}
\newcommand{\best}[1]{\textbf{#1}}

\newcommand{\poschg}[1]{\textcolor{green!45!black}{#1}}
\newcommand{\negchg}[1]{\textcolor{red!55!black}{#1}}

\newcommand{\artifacturl}{https://huggingface.co/datasets/Randallhy/RefineCut-Bench}
\newcommand{\codeurl}{https://github.com/Lancelot-wy/RefineCut}
\newcommand{\ghA}[1]{\textbf{\textcolor{pbBluetitle}{#1}}}
\newcommand{\ghB}[1]{\textbf{\textcolor{amberhi}{#1}}}

\definecolor{promptbg}{RGB}{248,248,248}
\definecolor{promptframe}{RGB}{210,210,210}
\definecolor{pbBluetitle}{RGB}{ 70,118,178}
\definecolor{pbBluebg}{RGB}{226,236,250}
\definecolor{pbPinktitle}{RGB}{188, 87,108}
\definecolor{pbPinkbg}{RGB}{250,228,232}
\definecolor{pbPurpletitle}{RGB}{124, 92,178}
\definecolor{pbPurplebg}{RGB}{236,228,250}
\definecolor{pbOrangetitle}{RGB}{200,124, 44}
\definecolor{pbOrangebg}{RGB}{252,236,212}
\definecolor{pbGreentitle}{RGB}{ 70,140, 90}
\definecolor{pbGreenbg}{RGB}{226,244,230}
\definecolor{pbTealtitle}{RGB}{ 50,130,130}
\definecolor{pbTealbg}{RGB}{222,242,242}

\tcbset{
  promptbase/.style={
    enhanced,
    width=\columnwidth,
    arc=0.8mm, outer arc=0.8mm,
    boxrule=0.4pt,
    left=1.5mm, right=1.5mm, top=0.9mm, bottom=0.9mm,
    toptitle=0.5mm, bottomtitle=0.5mm,
    fonttitle=\bfseries\small\color{white},
    coltitle=white,
    before skip=7pt, after skip=8pt,
    listing only,
    listing options={
      basicstyle=\ttfamily\scriptsize,
      breaklines=true, columns=fullflexible,
      keepspaces=true, showstringspaces=false
    },
  },
  pbneutral/.style={promptbase,
    colback=promptbg, colframe=promptframe, colbacktitle=black!8,
    fonttitle=\bfseries\footnotesize\color{black!80}, coltitle=black!80},
  pbblue/.style={pbneutral},
  pbpink/.style={pbneutral},
  pbpurple/.style={pbneutral},
  pborange/.style={pbneutral},
  pbgreen/.style={pbneutral},
  pbteal/.style={pbneutral},
}
\newtcblisting{promptbox}[2][pbblue]{
  #1, title={#2},
}

\tcbset{
  schemabase/.style={
    breakable, enhanced,
    width=\columnwidth,
    arc=0.4mm, outer arc=0.4mm,
    boxrule=0.4pt,
    left=0.9mm, right=0.9mm, top=0.7mm, bottom=0.7mm,
    toptitle=0.45mm, bottomtitle=0.45mm,
    fonttitle=\bfseries\small\ttfamily\color{white},
    coltitle=white,
    before skip=7pt, after skip=9pt,
  },
  sbblue/.style={schemabase, colback=pbBluebg!55, colframe=pbBluetitle,
    colbacktitle=pbBluetitle},
  sbpink/.style={schemabase, colback=pbPinkbg!55, colframe=pbPinktitle,
    colbacktitle=pbPinktitle},
  sbpurple/.style={schemabase, colback=pbPurplebg!55, colframe=pbPurpletitle,
    colbacktitle=pbPurpletitle},
  sborange/.style={schemabase, colback=pbOrangebg!55, colframe=pbOrangetitle,
    colbacktitle=pbOrangetitle},
  sbgreen/.style={schemabase, colback=pbGreenbg!55, colframe=pbGreentitle,
    colbacktitle=pbGreentitle},
  sbteal/.style={schemabase, colback=pbTealbg!55, colframe=pbTealtitle,
    colbacktitle=pbTealtitle},
}
\newtcolorbox{schemabox}[2][sbblue]{#1, title={#2}}

\usepackage{listings}
\AtBeginDocument{%
  \ifdefined\linenumbers
    \BeforeBeginEnvironment{lstlisting}{\par\nolinenumbers}%
    \AfterEndEnvironment{lstlisting}{\linenumbers}%
  \fi
}
\lstdefinelanguage{json}{
    basicstyle=\scriptsize\ttfamily,
    breaklines=true,
    breakatwhitespace=true,
    columns=fullflexible,
    showstringspaces=false,
    keepspaces=true,
    keywords={true,false,null},
    keywordstyle=\color{deltadn},
    stringstyle=\color{deltaup!70!black},
    morestring=[b]",
    frame=single,
    framerule=0pt,
    backgroundcolor=\color{bandgray!60},
    xleftmargin=2pt,
    xrightmargin=2pt,
    aboveskip=12pt,
    belowskip=14pt,
}
\lstdefinelanguage{jsonbare}{
    basicstyle=\scriptsize\ttfamily,
    breaklines=true,
    breakatwhitespace=true,
    columns=fullflexible,
    showstringspaces=false,
    keepspaces=true,
    keywords={true,false,null},
    keywordstyle=\color{deltadn},
    stringstyle=\color{deltaup!70!black},
    morestring=[b]",
    frame=none,
    xleftmargin=0pt,
    xrightmargin=0pt,
    aboveskip=0pt,
    belowskip=0pt,
}
\lstdefinelanguage{pyish}{
    basicstyle=\scriptsize\ttfamily,
    breaklines=true,
    columns=fullflexible,
    showstringspaces=false,
    keepspaces=true,
    keywords={def,for,in,if,else,elif,return,not,and,or,True,False,None,assert,while,break,continue},
    keywordstyle=\bfseries\color{deltadn},
    commentstyle=\color{gray!80!black}\itshape,
    stringstyle=\color{deltaup!70!black},
    morecomment=[l]{\#},
    morestring=[b]",
    frame=single,
    framerule=0pt,
    backgroundcolor=\color{bandgray!60},
    xleftmargin=2pt,
    xrightmargin=2pt,
    aboveskip=12pt,
    belowskip=14pt,
}

\title{Plans You Can Check: Verifier-Grounded Learning \\
of an Open-Weight Planner for Executable Video-Editing}

\author{
  \textbf{Haoyu Wang}\textsuperscript{1,2,\ensuremath{\dagger}},
  \textbf{Cheng Feng}\textsuperscript{3,\ensuremath{\dagger}},
  \textbf{Liuyang Bian}\textsuperscript{2,\ensuremath{\dagger}},
  \textbf{Ruiyang Huang}\textsuperscript{4,5,\ensuremath{\dagger}} \\
  \textbf{Lei Wei}\textsuperscript{5},
  \textbf{Yafei Wen}\textsuperscript{2},
  \textbf{Xiaoxin Chen}\textsuperscript{2},
  \textbf{Xiaoying Tang}\textsuperscript{6,7,8,\faEnvelopeO} \\
  \textsuperscript{1}School of Artificial Intelligence, The Chinese University of Hong Kong, Shenzhen \\
  \textsuperscript{2}vivo AI Lab \quad
  \textsuperscript{3}University of the Chinese Academy of Sciences \\
  \textsuperscript{4}Southeast University \quad
  \textsuperscript{5}Peking University \\
  \textsuperscript{6}School of Science and Engineering, The Chinese University of Hong Kong, Shenzhen \\
  \textsuperscript{7}Shenzhen Future Network of Intelligence Institute (FNii-Shenzhen) \\
  \textsuperscript{8}Guangdong Provincial Key Laboratory of Future Networks of Intelligence, CUHK-Shenzhen \\
  {\small
    \textsuperscript{\faEnvelopeO}\href{mailto:tangxiaoying@cuhk.edu.cn}{\texttt{tangxiaoying@cuhk.edu.cn}}
  }
}

\begin{document}
\maketitle

\begingroup
\renewcommand{\theHfootnote}{authorroles}
\let\svthefootnote\thefootnote
\let\thefootnote\relax
\footnotetext[0]{\textsuperscript{$\dagger$}\ Equal contribution.\quad
  \textsuperscript{\faEnvelopeO}\ Corresponding author.}
\let\thefootnote\svthefootnote
\endgroup

\begin{abstract}
Practical video editing is not only pixel generation: an editor
must turn a brief, a clip pool, music metadata, and hard
constraints into an executable timeline. We study this decision
layer as \emph{executable video-editing planning} and introduce
RefineCut, which, unlike workflow systems that wrap a prompted
frontier model, trains a compact open-weight planner for it.
The planner edits a typed timeline through structured patches
covering clip selection, trimming, ordering, transitions, and
duration and music alignment; a deterministic verifier applies
each patch and checks it against an explicit constraint ledger.
Because editing has no single ground-truth repair, we do not
imitate teachers directly: RefineCut replays every multi-teacher
branch through the verifier and keeps verifier-best repairs as
supervision. A second stage, RefineCut-Evo, lets the student score
its own repairs with the verifier and a task rubric and trains on
high-margin preference pairs, so the final $8$B planner runs in a
closed verifier loop with no teacher calls at inference. On
RefineCut-Bench ($3{,}578$ tasks, $7{,}971$ captioned clips, $499$
music tracks, explicit ledgers), verifier-replayed distillation
lifts the planner from $0.620$ to $0.858$ on the protocol-specific
Video-Editing Score and RefineCut-Evo reaches $0.924$; the gain
transfers to Llama-3.1-8B and GLM-4-9B, and in the same closed
loop the $8$B planner matches or exceeds its frontier teachers.
Code and RefineCut-Bench are publicly released; see the
\hyperref[sec:availability]{Data Availability} statement.
\end{abstract}

\begin{figure}[t]
  \centering
  \includegraphics[width=\columnwidth]{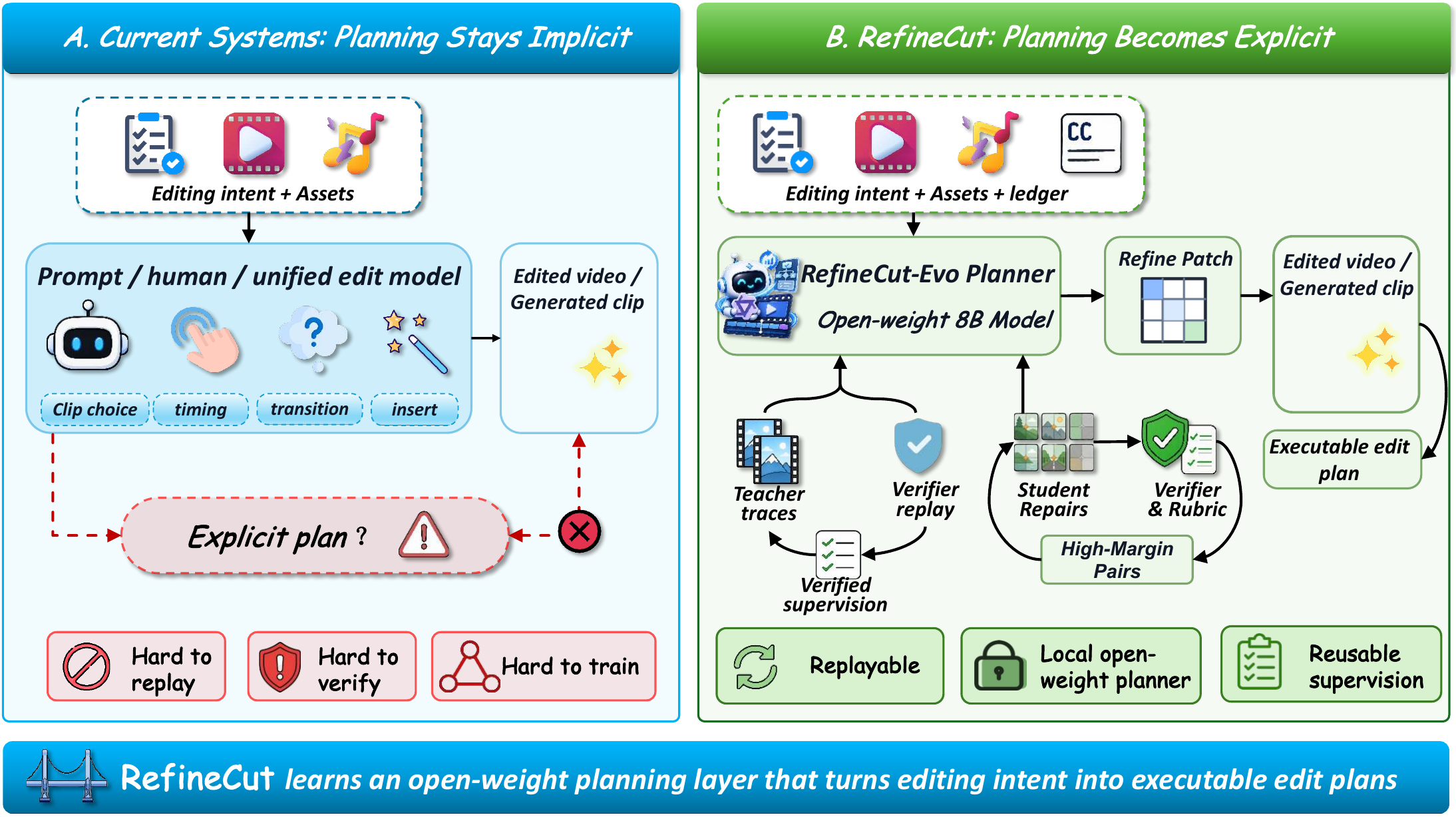}
  \caption{\textbf{From implicit to explicit planning.}
  RefineCut turns implicit prompted editing into explicit,
  verifier-trained executable planning.}
  \label{fig:teaser}
\end{figure}

\section{Introduction}
\label{sec:intro}

Real-world video editing is rarely about generating new pixels.
Given a pool of existing footage, an editor decides which clips
to keep, how to trim and order them, which transitions to use,
and how to align cuts to music. These decisions are tightly
constrained: the final cut must hit a target duration, retain or
exclude specific clips, respect pacing, and stay synchronized to
a soundtrack. The deliverable is a structured \emph{edit plan},
executed afterwards by a downstream toolchain. Editing, in other
words, is first a \emph{constrained planning} problem and only
later a rendering problem.

Recent systems address the layer below this decision process or
sidestep it entirely. Text-to-video and image-to-video models
generate or transform pixels directly
\citep{zhuang2024vlogger,long2024videostudio}, and
instruction-editing frameworks modify content within a single
shot \citep{yoon-etal-2025-raccoon}; none produces a clip-level
plan over a real footage pool. Workflow-oriented mashup and
dubbing systems
\citep{li2026direct,zhang2025lvasagent,hu2024storyagent,lin2026glance,liang2025univa}
do operate at the decision layer, but realize it by prompting a
frozen frontier backbone inside a hand-built scaffold. The cost
is real: the planning policy is never updated, so it cannot be
optimized against the hard constraints a brief imposes, and is
locked to a closed-source model. What the community lacks is a
compact, open-weight planner that learns editing decisions and
can sit in front of any rendering back-end, as shown in
Figure~\ref{fig:teaser}.

We argue this gap is also a missed opportunity, because the
editing decision layer has a property open-ended generation
lacks: its outputs can be checked. We formalize the layer
as \emph{executable video-editing planning}. A task instance
gives the planner a brief, a real clip pool with captions and
metadata, optional music with beat tracks, the current timeline
state, and, crucially, an explicit \emph{constraint ledger}
enumerating what the final cut must satisfy. The planner emits a
\emph{RefinePatch}: an RFC~6902-style JSON Patch~\citep{rfc6902}
over a typed timeline state. Because every requirement is an
explicit ledger entry, a \emph{deterministic verifier}, rather
than a learned LLM judge \citep{zheng2023judging}, can apply
each patch and recompute the ledger entry by entry, following
execution-based supervision in verifiable domains such as
mathematical reasoning and code
\citep{cobbe2021gsm8k,lightman2024verify,lambert2024tulu3}. This
verifiability is what makes the planning layer learnable in a way
pixel-level generation is not.

Turning this observation into a trained planner is still not a
single-target imitation problem. The natural starting point is
to distill trajectories from frontier models, since tool-use
trajectories can supervise planner behavior
\citep{zeng-etal-2024-agenttuning} and step-level feedback can
turn successful and failed attempts into preference signals
\citep{song2024eto,xiong2024watch}. But editing has no single
ground-truth repair: the same brief admits many valid cuts, so a
teacher's first-choice branch is a guess, not a label, which is
why preference-based learning, not imitation of one reference,
is standard for open-ended generation
\citep{christiano2017preferences,stiennon2020summarize,ouyang2022instructgpt}.
A planner that imitates noisy teacher traces inherits their
noise. We therefore replay every candidate branch through the
verifier before using it, and let the student keep improving on
its own verified repairs after distillation
\citep{madaan2023selfrefine,yuan2024selfrewarding}.

RefineCut is a two-stage framework built on this logic: editing
decisions should be learned and verified, not prompted. Stage~1,
the verifier-replayed teacher bootstrap, canonicalizes traces
from three frontier API teachers, namely
GPT-5.4\footnote{GPT-5.4: accessed via API in 2026, no public
technical report at the time of writing. Qwen3-Max:
\url{https://huggingface.co/Qwen}.},
Qwen3-Max, and
DeepSeek-V4-Pro~\citep{deepseekai2026deepseekv4}, replays each
candidate branch through the verifier, and keeps verifier-best
branches as SFT targets and mixed-granularity preference pairs,
yielding the Mixed-Pref 8B planner. Stage~2, RefineCut-Evo,
performs verifier-centered, rubric-structured self-improvement:
the student scores its own repair candidates with the verifier
and a task-specific rubric (ER1--ER7), and high-margin pairs
train it with Direct Preference Optimization
(DPO)~\citep{rafailov2023dpo}. Supervision thus shifts from
external teachers to verifier-grounded self-improvement, and at
inference the planner runs in a closed verifier loop with no
teacher calls.

\paragraph{Contributions.}
\begin{itemize}
\setlength\itemsep{1pt}
\item We formulate executable video-editing planning and
release RefineCut-Bench, a planning-level benchmark with real
clip/music metadata, constraint ledgers, and multi-teacher
trajectories.
\item We propose \emph{verifier-replayed trajectory distillation}:
canonical normalization plus deterministic replay that converts
noisy frontier traces into verified SFT targets and
mixed-granularity preference pairs.
\item We introduce RefineCut-Evo, an
EvoLM-inspired~\citep{li2026evolmselfevolvinglanguagemodels}
verifier-centered, rubric-structured self-improvement stage
trained with DPO on verifier- and rubric-scored candidates.
\item Across Qwen3-8B, Llama-3.1-8B, and GLM-4-9B, verifier
replay beats raw imitation, and the final $8$B planner matches
its frontier teachers in the same closed loop
(Section~\ref{sec:experiments}).
\end{itemize}

\paragraph{Positioning.}
RefineCut is closest to workflow-oriented video and multimodal
creation systems such as DIRECT, LVAS-Agent, GLANCE, StoryAgent,
and UniVA
\citep{li2026direct,zhang2025lvasagent,lin2026glance,hu2024storyagent,liang2025univa},
but those systems keep the editing policy inside prompted
frontier-model workflows. It also builds on tool-executable
planner learning and on self-improvement ideas from EvoLM and
Rubric-Grounded RL
\citep{li2026evolmselfevolvinglanguagemodels,bhattarai2026rubricgroundedrlstructuredjudge},
while using a deterministic editing verifier rather than a
general LLM judge as the primary training signal.
Appendix~\ref{app:related} gives the full related-work
discussion.

\section{Task and Benchmark}
\label{sec:task}

This section formalizes planning-level video editing as a task,
derives the evaluation principles used throughout the paper, and
introduces RefineCut-Bench as the dataset, verifier, and
multi-teacher trajectory resource that Section~\ref{sec:method}
turns into supervision.

\subsection{Planning-Level Video Editing}
\label{sec:task-plan}

RefineCut studies the decision layer of video editing: the
layer that turns an editing intent and a pool of footage into an
executable edit plan. The final video is rendered by a
downstream toolchain, not by the planner, which never observes
pixels and reads only textual clip captions and visual metadata
from an upstream vision-language captioner, as in prior work
where a language model reasons over captions rather than raw
frames \citep{zeng2022socratic,krishna2017dense}.

\paragraph{Task definition.}
A task instance is a tuple
\begin{equation}
x = (b,\; C,\; M,\; s_0,\; L),
\label{eq:task-tuple}
\end{equation}
where $b$ is a natural-language brief; $C$ is a real clip pool,
each clip carrying a caption and visual metadata (scene, motion
intensity, camera duration); $M$ is optional music metadata
(beat times and energy); $s_0$ is the initial timeline state; and
$L$ is an explicit \emph{constraint ledger} enumerating what the
final cut must satisfy. A planner is a policy $\pi_\theta$ that at
step $t$ emits an edit patch
$p_t \sim \pi_\theta(\cdot \mid x, s_t, L_t)$; a deterministic
verifier applies it, advancing the state as
$s_{t+1} = \mathrm{Apply}(s_t, p_t)$ and recomputing the ledger
as $L_{t+1} = \mathrm{Verify}(s_{t+1})$. The objective is to
reach a ledger-satisfying state within a small repair budget; the
closed-loop execution is detailed in Section~\ref{sec:method}.

\subsection{Inputs, Outputs, and Editing State}
\label{sec:task-io}

\paragraph{Clip pool and metadata.}
Clip captions and visual metadata are produced offline by an
upstream vision-language captioner
\citep{krishna2017dense,wang2024qwen2vl}; music metadata (beat
times, energy) is extracted by a standard beat-tracking pipeline
and supports music-synchronization constraints
\citep{davis2018visual}.

\paragraph{Editing state and edit artifact.}
The state $s_t$ holds the selected clip sequence, segment
durations, transition styles, music-sync parameters, generated
placeholder slots, and per-entry ledger satisfaction. The planner
edits it through a RefinePatch, an RFC~6902-style JSON
Patch over the typed timeline state; a typed
patch (rather than free-form text) is what makes each step
machine-applicable and machine-checkable. When the brief calls
for content absent from the pool, a placeholder slot records the
requested prompt and position, with visual generation delegated
downstream. Prompts use task-local clip aliases, so measured
performance reflects planning ability rather than memorization of
global clip identifiers. Schema details are in
Appendix~\ref{app:schemas}.

\paragraph{Constraint ledger $L$.}
Each ledger entry is a tuple
$(\textit{item\_id},\, \textit{type},\, \textit{spec},\,
\textit{satisfied},\, \textit{evidence})$, spanning seven
field-level constraint families: duration, transition,
music\_sync, clip inclusion, clip exclusion, repeat limit, and
pacing; generated-placeholder requirements are represented as
task-structure constraints checked by the same verifier (subtypes
in Appendix~\ref{app:bench-ledger}). The
ledger makes editing requirements explicit and machine-checkable,
as in verifiable instruction-following
\citep{zhou2023instruction,jiang2024followbench}. An entry is a
\emph{hard constraint} when its \textit{spec} admits a
deterministic pass/fail test (e.g., a target duration within
tolerance, a clip that must be kept or excluded), and
\textsc{HardPass} is the event that every hard constraint holds
simultaneously. Softer entries earn graded credit: keeping two
of three required clips fails the \texttt{must\_keep\_clip} entry
(and \textsc{HardPass}) but still counts toward the
constraint-satisfaction rate and a required-clip recall of
$2/3$. The verifier, not the planner, recomputes the
\textit{satisfied} field after every patch.

\subsection{What Makes an Editing Plan Valid?}
\label{sec:task-validity}

\begin{figure*}[!t]
\centering
\includegraphics[width=0.96\textwidth]{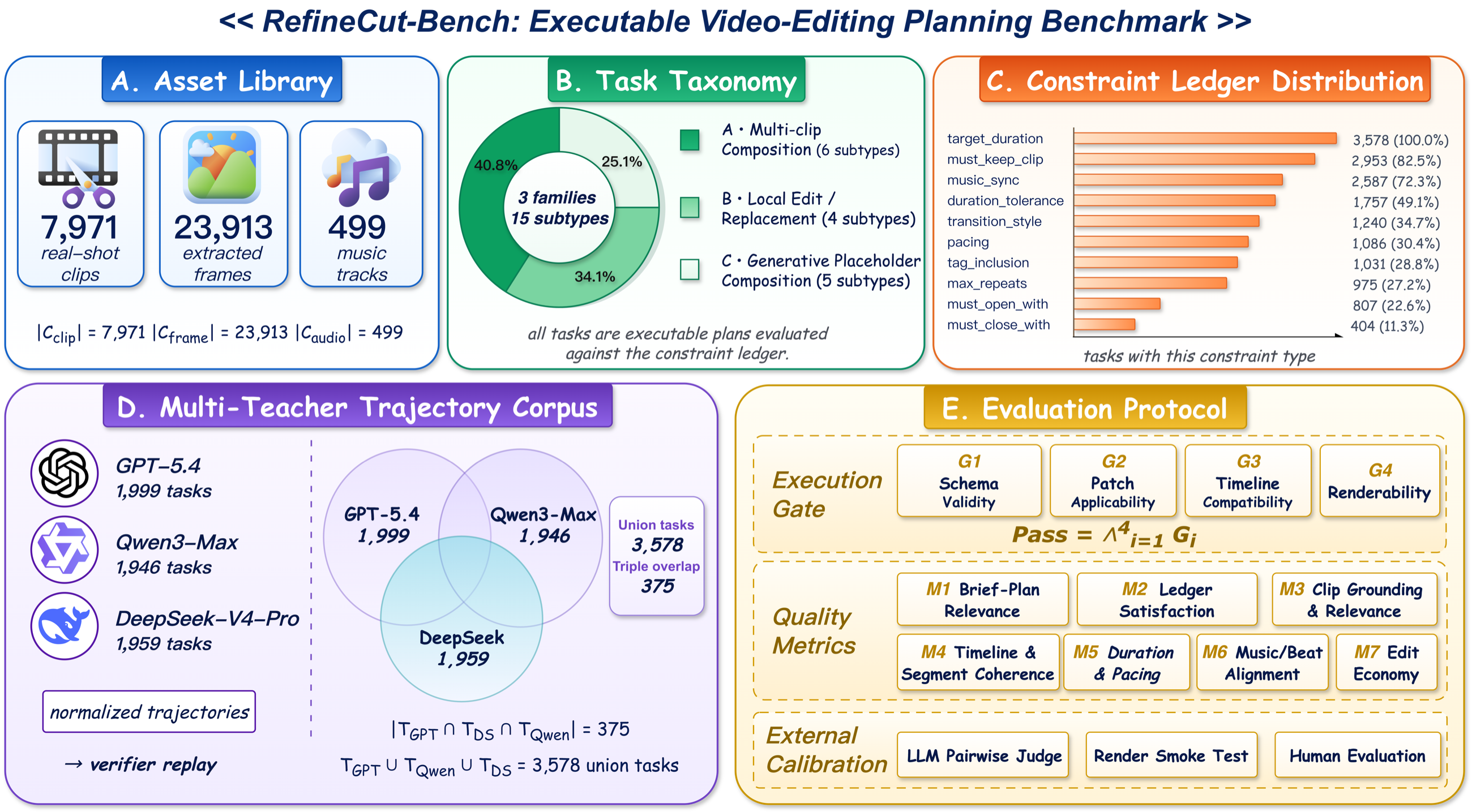}
\caption{\textbf{RefineCut-Bench overview.}
The benchmark ties real clip/music metadata, explicit ledger
constraints, multi-teacher trajectories, and verifier-based
evaluation into one planning-level protocol.}
\label{fig:bench}
\end{figure*}

A valid edit plan must be executable, grounded in the given
clips, satisfy the explicit constraints without breaking
satisfied ones, and converge within a small repair budget. These
requirements form a hierarchy, which our evaluation mirrors with
three layers. The first layer, the execution gate, is a hard
prerequisite covering schema validity, patch applicability, and
timeline validity; a plan that fails it is not executable at
all, mirroring execution-based evaluation in code generation
\citep{jimenez2024swebench}. Conditioned on
the gate, the second layer, planning quality, asks whether the
executable plan actually satisfies the brief, from constraint
satisfaction and clip grounding to duration, pacing, and
convergence. Both layers are decided entirely
by the deterministic verifier, so every score is a checkable
consequence of the ledger rather than a learned judgment.
Finally, the third layer, human-rendered quality, is assessed
after rendering and covers instruction fit, content relevance,
temporal coherence, shot continuity, pacing, audio-visual rhythm,
and overall quality, following established human-evaluation
practice for edited video
\citep{huang2024vbench,liu2024evalcrafter}. The two families are
complementary: verifier metrics test whether a plan executes
under the ledger, human evaluation whether the rendered edit is
perceptually preferred. We report component metrics with the
\textsc{VES} summary; definitions and weights are in
Section~\ref{sec:exp-metrics} and Appendix~\ref{app:metrics}.

\subsection{RefineCut-Bench}
\label{sec:bench}

RefineCut-Bench instantiates the task above as a controlled,
executable-planning benchmark. Existing video-editing resources
either evaluate rendered pixels or wrap a prompted frontier model
in an agent scaffold whose decisions are never checked against an
explicit specification; to our knowledge, RefineCut-Bench is the
first to tie four ingredients into one planning-level release:
real clip and music metadata, explicit constraint ledgers,
multi-teacher trajectories, and a deterministic verifier-based
evaluation protocol (Appendix~\ref{app:bench-details},
Figure~\ref{fig:bench}, Table~\ref{tab:split}).

\paragraph{Assets, tasks, and splits.}
The asset library draws from $7{,}971$ captioned clips,
$23{,}913$ caption-anchored frames, and $499$ music tracks. Tasks
span three families, (A)~multi-clip composition, (B)~targeted
insertion or replacement, and (C)~generative composition with
placeholder slots, totaling $15$ subtypes and $3{,}578$ canonical
tasks. Raw records are deduplicated into canonical tasks and
split at the record level. Because resampled variants of one task
can otherwise straddle the train/test boundary, we mark a
\emph{canonical-clean} test subset whose canonical identifiers
never appear in training, and use it to check that headline
gains cannot be attributed to canonical-id overlap
(Section~\ref{sec:exp-analysis}). Split
sizes, per-subtype and per-constraint distributions, and the
captioning pipeline are in Appendix~\ref{app:bench-details} and
Table~\ref{tab:split}.

\subsection{Multi-Teacher Trajectories as Noisy Supervision}
\label{sec:bench-traj}

RefineCut-Bench also releases a multi-teacher trajectory
resource, like recent releases of agent trajectories as a
training signal rather than only an evaluation target
\citep{zeng-etal-2024-agenttuning,song2024agentbank}. For every
training task we run one refinement rollout from each of three
independent frontier API teachers (GPT-5.4, Qwen3-Max, and
DeepSeek-V4-Pro); at each refine step a teacher emits several
candidate branches, each a proposed RefinePatch, with its
own first-choice flag (coverage and schema in
Appendix~\ref{app:bench-traj}).

This resource is noisy by construction. Because editing
admits no single ground-truth repair, teachers disagree on
schema, branch structure, and which branch is best; a
first-choice flag is a guess, not a label, and even one teacher's
branches vary in executable quality. Multi-teacher distillation
that imitates such heterogeneous demonstrations directly would
inherit their inconsistency \citep{tian2024tinyllm}. The
constraint ledger offers a way out: every branch can be
canonicalized and replayed through the deterministic verifier,
which scores it against the same explicit ledger regardless of
its source. Verifier replay thus converts an inherently noisy
resource into a consistently graded supervision signal, the
mechanism Section~\ref{sec:method} builds on.

\section{Method: Verifier-Replayed Distillation and Verifier-Centered Self-Improvement}
\label{sec:method}

\begin{figure*}[!t]
  \centering
  \includegraphics[width=\textwidth]{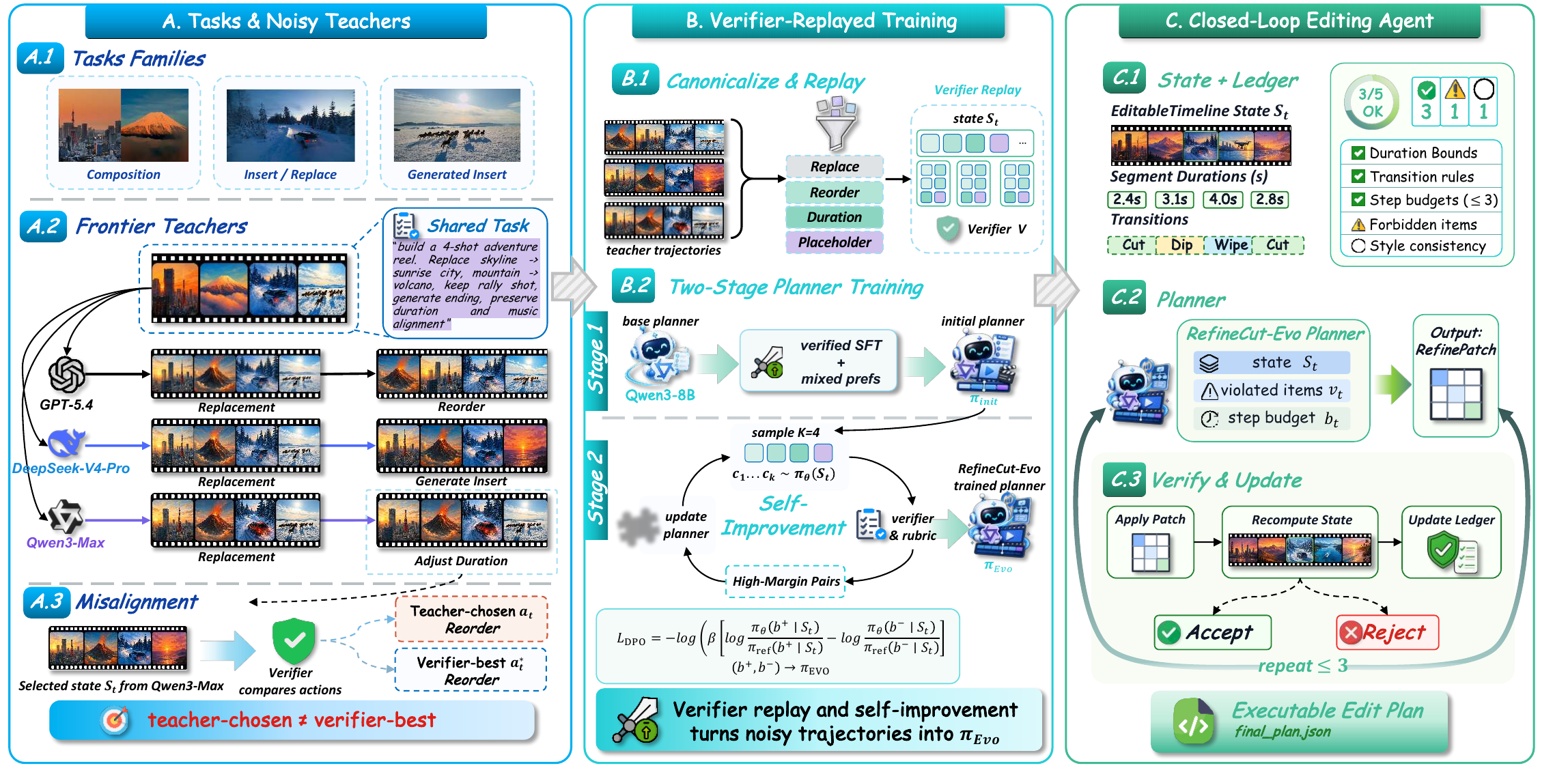}
  \caption{\textbf{Framework.}
  Noisy teacher trajectories become verified supervision; verifier- and rubric-scored
  student repairs drive self-improvement and closed-loop planning.}
  \label{fig:framework}
\end{figure*}

\subsection{Overview and Trajectory Collection}
\label{sec:method-overview}

Our method has two offline learning stages and one runtime loop,
shown in Figure~\ref{fig:framework}. Stage~1, verifier-replayed
teacher bootstrap, canonicalizes raw multi-teacher trajectories
and replays every candidate branch through the verifier;
verifier-best branches yield verified SFT targets and
mixed-granularity preference pairs that produce an initial 8B
planner, Mixed-Pref. Stage~2, RefineCut-Evo, samples $K{=}4$
student repairs on training states, scores them by verifier and
rubric ER1--ER7, and trains with DPO on high-margin pairs. At
runtime, the planner and verifier run the Apply/Verify closed
loop defined in Section~\ref{sec:task-plan} for at most $T{=}3$
steps.

Each teacher-covered task receives one refinement rollout from
each available frontier API teacher (GPT-5.4, Qwen3-Max, and
DeepSeek-V4-Pro). Each refine step contains multiple candidate
RefinePatch branches. The teacher's first-choice flag is recorded
but not treated as ground truth: the verifier picks the training
target. Per-teacher counts and the branch schema are in
Appendix~\ref{app:bench-traj}.

\subsection{Canonical Trajectory Normalization}
\label{sec:method-canonical}

Raw teacher outputs are heterogeneous: some wrap the task in an
envelope, some flatten multi-step trajectories, and others nest
branches as lists. We rewrite them into canonical patch
trajectories keyed by (\texttt{task\_id}, \texttt{teacher\_id}),
map JSON Pointer paths to a canonical
namespace, and validate clip references against task-local aliases.
Schema, path-alias table, and validation rules are in
Appendix~\ref{app:schemas}.

\subsection{Verifier Replay and Branch Arbitration}
\label{sec:method-replay}

For every canonicalized step we keep up to four branches with
distinct \texttt{repair\_operator}s and Jaccard-distinct clip
selections, then replay each through the deterministic verifier.
For branch $b$ at state $s_t$, the verifier validates and applies
the patch, recomputes the ledger $L'$, and scores six signals:
constraint-satisfaction change, targeted repair, required-clip
recall, patch applicability, no-regression, and locality. The
exact weighted definition is in Appendix~\ref{app:metrics}; we
write the scalar score as $V(b)$. The verifier-best branch
advances the state, and branch score gaps drive preference
pairing. The full procedure is summarized in
Appendix~\ref{app:rubric}, Algorithm~\ref{alg:replay}.

\subsection{Verifier-Replayed Distillation}
\label{sec:method-distillation}

Replay gives two supervision signals. Verified SFT retains the
verifier-best branch as $p^\star$ when it canonicalizes and
applies, repairs at least one violated ledger entry, regresses no
satisfied constraint, and, for clip-grounding repairs, references
a legitimate task-local alias. We fine-tune on these canonicalized
patches with standard next-token cross-entropy. Mixed-granularity
preference combines step-level pairs, in which the verifier-best
branch is paired against a low-scoring branch separated by margin
$\delta$, with trajectory-level pairs ranked by terminal verifier
score. We fine-tune the SFT checkpoint with offline
DPO~\citep{rafailov2023dpo}:
\begin{equation}
\begin{aligned}
\mathcal{L}_{\mathrm{DPO}}
&=-\mathbb{E}_{(x,p^+,p^-)}
\log\sigma\!\left(\beta z\right),\\
z&=\Delta_\theta(p^+,x)-\Delta_\theta(p^-,x),\\
\Delta_\theta(p,x)
&=\log\pi_\theta(p\mid x)-\log\pi_{\mathrm{ref}}(p\mid x).
\end{aligned}
\label{eq:dpo}
\end{equation}
Here $p^+$ and $p^-$ are the preferred and rejected branches,
$\pi_{\mathrm{ref}}$ is the frozen SFT checkpoint, and $\beta$
controls deviation from it. Preferences come from executable
verifier replay rather than LLM judgments; the resulting checkpoint
is Mixed-Pref.

\subsection{RefineCut-Evo: Verifier-Centered Student Self-Improvement}
\label{sec:method-evo}

RefineCut-Evo improves the student with its own repairs scored by
the verifier and a fixed, task-specific editing rubric, following
recent rubric-guided self-improvement work
\citep{li2026evolmselfevolvinglanguagemodels,bhattarai2026rubricgroundedrlstructuredjudge}.

\paragraph{Candidates and rubric.}
For each training state we sample $K{=}4$ RefinePatches from
Mixed-Pref, parse and replay them, and filter schema or apply
failures. The rubric ER1--ER7 covers Intent, Ledger
Satisfaction, Clip Grounding, Timeline Coherence,
Duration/Pacing, Music/Beat, and Edit Economy;
per-family weights and deterministic fallback proxies are in
Appendix~\ref{app:rubric}.

\paragraph{Joint scoring and training.}
Writing $R(c)=\sum_{m=1}^{7}\alpha_m r_m(c)$ with
$\sum_m\alpha_m=1$, each candidate receives
$S(c)=\lambda V(c)+(1-\lambda)R(c)$, where $V(c)$ is the verifier
score from Section~\ref{sec:method-replay} and $\lambda=0.65$.
At each state we choose
\begin{equation}
\begin{aligned}
c^+ &= \arg\max_{c\in\mathcal{C}(s)}S(c),\\
c^- &\in \{c\in\mathcal{C}(s):S(c^+)-S(c)\ge\tau\}.
\end{aligned}
\label{eq:evo-pair}
\end{equation}
Thus $c^-$ is a lower-scoring hard negative rather than the lowest
sampled candidate. RefineCut-Evo starts from Mixed-Pref and trains
with DPO on these pairs; hyperparameters and the dev100 selection
rule are in Appendix~\ref{app:training}. Section~\ref{sec:exp-rq4}
analyzes the verifier and rubric contributions.

\subsection{Closed-loop Deployment}
\label{sec:method-loop}

At test time the planner reads $s_t$ and the still-violated ledger
entries and emits one RefinePatch; the verifier validates, applies,
and recomputes the ledger. No teacher is called at inference. The
loop repeats for at most $T{=}3$ steps, after
which the final state is dispatched to the editing toolchain.

\section{Experiments}
\label{sec:experiments}

We address five research questions (RQ1--RQ5), a robustness and
failure analysis, and a blind rendered-preview evaluation.

\subsection{Experimental Setup}
\label{sec:exp-setup}

\textbf{Benchmark.} All main results are on Common-100, with a
single PatchPlanner prompt shared by every model;
canonical-clean ($N{=}92$) is held out for robustness checks.
dev100 is used once, to pick the RefineCut-Evo checkpoint
(step~$300$; Appendix~\ref{app:additional},
Table~\ref{tab:dev-selection}), before any test-set measurement. At test time the
planner interacts only with the verifier; no teacher is called.

\textbf{Loop and budget.} The closed loop runs at most $T{=}3$
repair steps, the same horizon used to collect teacher
trajectories (Appendix~\ref{app:bench-traj}) and to define
Converged@3. During Evo training we draw $K{=}4$ student
candidates per state, matching the teacher branch budget.

\textbf{Models.} The main progression trains Qwen3-8B-Instruct
with LoRA~\citep{hu2022lora}: Prompted (no training, same
verifier loop), Raw, Verified, Traj-Pref, Mixed-Pref, and
RefineCut-Evo, all built on the teacher trajectories of
Section~\ref{sec:bench-traj}. To check that verifier-replayed
supervision is not tied to one backbone, we rerun the
Prompted\,/\,Raw\,/\,Verified comparison on Llama-3.1-8B-Instruct
and GLM-4-9B with identical data, recipe, and protocol
(Section~\ref{sec:exp-rq5}). Hyperparameters and prompts are in
Appendices~\ref{app:training}--\ref{app:prompts}.

\subsection{Evaluation Metrics}
\label{sec:exp-metrics}

Automatic metrics evaluate executable planning: schema validity,
patch applicability, timeline validity, constraint satisfaction,
hard-constraint pass, clip grounding, duration/pacing,
no-regression, and convergence. The Video-Editing Score
(\textsc{VES}) aggregates them as
\begin{equation*}
\setlength{\jot}{1pt}
\small
\begin{aligned}
\mathrm{VES}={}&0.30\,\mathrm{FinalCSR}+0.15\,\mathrm{HardPass}\\
&+0.15\,\mathrm{PASR}+0.15\,\mathrm{ReqClipRecall}\\
&+0.10\,\mathrm{DurationPass}+0.10\,\mathrm{TimelineValidity}\\
&+0.05\,\mathrm{NoRegression}.
\end{aligned}
\end{equation*}
The same frozen ledger and verifier that score \textsc{VES} also
drive replay and closed-loop feedback, so \textsc{VES} is an
in-protocol measure, not an independent judgment; we always
report the component metrics with it and check rendered-preview
preference separately (Section~\ref{sec:exp-human}). Definitions
and weight sensitivity are in
Appendices~\ref{app:metrics}--\ref{app:additional}; tables
abbreviate Converged@3 as \textsc{Cvg}.

\subsection{RQ1: Can an Open-Weight Planner Learn Executable
Editing Planning?}
\label{sec:exp-rq1}

\begin{figure}[t]
\centering
\includegraphics[width=.72\columnwidth]{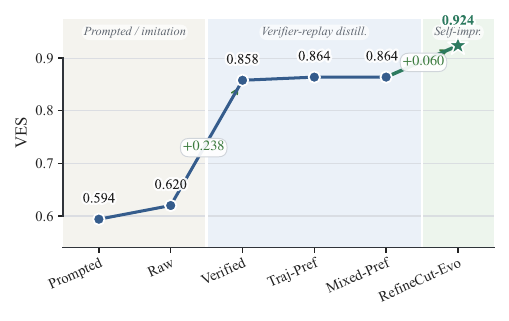}
\caption{\textbf{Training-stage progression.}
Common-100 VES across checkpoints; annotations mark the
Raw$\to$Verified and Mixed-Pref$\to$Evo aggregate gains.}
\label{fig:training-progression}
\end{figure}

An open-weight planner can learn executable editing planning,
and the two training stages contribute in different ways
(Table~\ref{tab:main}, Figure~\ref{fig:training-progression}).
Verifier-replayed SFT does most of the work, lifting \textsc{VES}
from $0.620$ to $0.858$; preference training over the same traces
adds little on aggregate ($0.864$) until RefineCut-Evo reaches
$0.924$. Where the Evo gain lands explains what it adds:
\textsc{HardPass} rises $0.670{\to}0.820$, \textsc{Dur@2s}
$0.830{\to}0.980$, and \textsc{Converged@3} $0.800{\to}0.950$ --
precisely the all-or-nothing checks a finished plan must pass,
not properties of any single patch. Distillation teaches the
planner to make edits that execute; Evo teaches it to finish
plans that satisfy the whole brief.

\begin{table}[t]
\centering
\scriptsize
\setlength{\tabcolsep}{1.8pt}
\renewcommand{\arraystretch}{1.02}
\begin{tabular*}{\columnwidth}{@{\extracolsep{\fill}}lrrrrrr@{}}
\toprule
& \multicolumn{3}{c}{\ghA{Constraint adherence}}
& \multicolumn{2}{c}{\ghB{Loop quality}}
& \multicolumn{1}{c}{\textbf{Overall}} \\
\cmidrule(lr){2-4}\cmidrule(lr){5-6}\cmidrule(l){7-7}
\textbf{Model}
& \multicolumn{1}{c}{\textbf{FCSR}}
& \multicolumn{1}{c}{\textbf{Hard}}
& \multicolumn{1}{c}{\textbf{ReqCR}}
& \multicolumn{1}{c}{\textbf{Dur}}
& \multicolumn{1}{c}{\textbf{Cvg}}
& \multicolumn{1}{c}{\textbf{VES}} \\
\midrule
\rowcolor{gray!6} \multicolumn{7}{@{}l}{\textit{Prompted backbone and raw imitation}} \\
Prompted              & 0.608 & 0.150 & 0.273 & 0.660 & 0.210 & 0.594 \\
Raw                   & 0.559 & 0.160 & 0.723 & 0.440 & 0.250 & 0.620 \\
\addlinespace[1pt]
\midrule
\rowcolor{blue!5} \multicolumn{7}{@{}l}{\textit{Verifier replay distillation}} \\
Verified              & 0.880 & 0.630 & 0.975 & 0.820 & 0.790 & 0.858 \\
Traj-Pref             & 0.877 & 0.670 & 0.969 & 0.840 & 0.790 & 0.864 \\
Mixed-Pref            & 0.882 & 0.670 & 0.969 & 0.830 & 0.800 & 0.864 \\
\addlinespace[1pt]
\midrule
\rowcolor{green!8} \multicolumn{7}{@{}l}{\textit{Self-improvement (final)}} \\
\rowcolor{green!8}
\textbf{RefineCut-Evo}$^{\star}$
	                      & \best{0.953} & \best{0.820} & \best{0.981}
	                      & \best{0.980} & \best{0.950} & \best{0.924} \\
\rowcolor{green!8}
\textit{$\Delta$ vs.\ Mixed-Pref}
	                      & \poschg{$+0.071$} & \poschg{$+0.150$} & \poschg{$+0.012$}
	                      & \poschg{$+0.150$} & \poschg{$+0.150$} & \poschg{$+0.060$} \\
\bottomrule
\end{tabular*}
\caption{\textbf{Closed-loop planning results on Common-100.}
The $\Delta$ row gives aggregate changes relative to Mixed-Pref;
$^{\star}$ marks the final RefineCut checkpoint.}
\label{tab:main}
\end{table}

\subsection{RQ2: Why is Verifier Replay Needed?}
\label{sec:exp-rq2}

Teacher traces cannot be imitated as they are. Each teacher
proposes several candidate branches per refine step, the three
teachers disagree on schema and branch structure, and even their
best branches are partial repairs rather than solved plans.
Before any branch becomes supervision, we therefore canonicalize
it, apply it to the real editing state, and let the verifier
score the result. Table~\ref{tab:teacher-main} shows what this
replay finds.

\begin{table}[t]
\centering
\scriptsize
\setlength{\tabcolsep}{2.2pt}
\renewcommand{\arraystretch}{1.02}
\begin{tabular*}{\columnwidth}{@{\extracolsep{\fill}}lrrrrrrr@{}}
\toprule
& \multicolumn{2}{c}{\ghA{Replay scale}}
& \multicolumn{5}{c}{\ghB{Verifier components}} \\
\cmidrule(lr){2-3}\cmidrule(l){4-8}
\textbf{Teacher}
& \multicolumn{1}{c}{\textbf{$N$}}
& \multicolumn{1}{c}{\textbf{Branches}}
& \multicolumn{1}{c}{\textbf{FCSR}}
& \multicolumn{1}{c}{\textbf{Hard}}
& \multicolumn{1}{c}{\textbf{PASR}}
& \multicolumn{1}{c}{\textbf{ReqCR}}
& \multicolumn{1}{c}{\textbf{VES}} \\
\midrule
GPT-5.4         & 2{,}000 & 6{,}000 & 0.379 & 0.015 & 0.872 & 0.698 & 0.521 \\
Qwen3-Max       & 1{,}946 & 3{,}892 & 0.446 & 0.016 & 0.989 & 0.712 & 0.557 \\
DeepSeek-V4-Pro & 1{,}959 & 3{,}920 & 0.473 & 0.028 & 0.946 & 0.703 & 0.559 \\
\bottomrule
\end{tabular*}
\caption{\textbf{Verifier-best replay diagnostics by teacher.}
Per-branch values are computed from replay logs; this diagnoses the
teacher resource, not student closed-loop performance.}
\label{tab:teacher-main}
\end{table}

Even the verifier-best branches, which almost always apply
cleanly (\textsc{PASR} $0.87$--$0.99$), almost never satisfy the
whole ledger in one step (\textsc{Hard} $\le 0.03$): a teacher's
first choice is a weak label, and which branch actually helps
depends on the state it lands on. Table~\ref{tab:main} shows the
consequence. Imitating teacher-selected branches yields
\textsc{VES} $0.620$, barely above the prompted backbone, while
training on verifier-best branches yields $0.858$ from a
training set three times smaller ($3{,}317$ vs.\ $9{,}690$
examples). The gain comes from replay filtering out branches
that fail on execution, not from more data.

\subsection{RQ3: Can RefineCut-Evo Self-Improve Beyond
Mixed-Pref?}
\label{sec:exp-rq3}

It does, and the improvement behaves like targeted repair rather
than a broad shift. Of the $100$ test tasks, Evo improves $22$
and degrades only $5$, leaving the other $73$ untouched, which
moves the aggregate from $0.864$ to $0.924$. The paired gain of
$+0.059$ VES (95\% CI $[0.028,0.092]$) is not noise, and the same
picture holds on canonical-clean and against the other distilled
variants (Appendix~\ref{app:additional}, Table~\ref{tab:paired}).

\subsection{RQ4: What Drives RefineCut-Evo?}
\label{sec:exp-rq4}

Evo differs from Mixed-Pref in two ingredients: preference pairs
are scored by the verifier, optionally refined by the rubric, and
only high-margin pairs are kept. Table~\ref{tab:ablation} removes
one ingredient at a time, holding backbone, candidate pool, DPO
recipe, and wall-clock budget fixed. Scoring pairs with the
verifier alone already reaches $0.909$, so certified repairs
carry most of the gain; rubric-structured margins add a further
$0.015$, concentrated in duration ($0.940{\to}0.980$) and
convergence ($0.900{\to}0.950$). Keeping every pair instead of
only high-margin ones costs $0.027$, and replacing the selection
scheme altogether with UCPO-lite~\citep{lochab2026ucpo} costs
$0.044$. Which pairs enter training matters more than how finely
each pair is scored.

\begin{table}[t]
\centering
\scriptsize
\setlength{\tabcolsep}{2.7pt}
\renewcommand{\arraystretch}{1.02}
\begin{tabular*}{\columnwidth}{@{\extracolsep{\fill}}lrrrrr@{}}
\toprule
& \multicolumn{4}{c}{\ghA{Component metrics}}
& \multicolumn{1}{c}{\textbf{Overall}} \\
\cmidrule(lr){2-5}\cmidrule(l){6-6}
\textbf{Variant}
& \multicolumn{1}{c}{\textbf{FCSR}}
& \multicolumn{1}{c}{\textbf{Hard}}
& \multicolumn{1}{c}{\textbf{Dur}}
& \multicolumn{1}{c}{\textbf{Cvg}}
& \multicolumn{1}{c}{\textbf{VES}} \\
\midrule
\rowcolor{green!8} Rubric-margin DPO & 0.9530 & 0.820 & 0.980 & 0.950 & \best{0.9237} \\
Verifier-only DPO & 0.9395 & 0.780 & 0.940 & 0.900 & 0.9085\dn{0.015} \\
No-margin DPO & 0.9178 & 0.760 & 0.910 & 0.870 & 0.8965\dn{0.027} \\
UCPO-lite DPO & 0.9012 & 0.740 & 0.890 & 0.840 & 0.8800\dn{0.044} \\
\bottomrule
\end{tabular*}
\caption{\textbf{RefineCut-Evo ablation} under matched
backbone, candidate pool, DPO recipe, and wall-clock budget.
Verifier-scored self-improvement is the primary signal;
rubric-structured margins add $0.015$.}
\label{tab:ablation}
\end{table}

\subsection{RQ5: Does Verifier Replay Transfer Across Backbones,
and Does RefineCut Compete with Frontier Policies?}
\label{sec:exp-rq5}

Verifier replay is not a Qwen-specific effect. Rerunning the
Prompted\,/\,Raw\,/\,Verified comparison on Llama-3.1-8B and
GLM-4-9B with the same data and recipe repeats the pattern
(Table~\ref{tab:backbone-main}): raw imitation is flat on Qwen
($+0.026$) and harmful on Llama and GLM ($-0.070$, $-0.035$),
while verified SFT recovers every family ($+0.238$, $+0.153$,
$+0.079$) and lifts required-clip recall to $0.98$, $0.92$, and
$0.99$. Absolute scores stay highest on Qwen3-8B, the only
backbone that also gets the preference stages.

\begin{table}[t]
\centering
\scriptsize
\setlength{\tabcolsep}{3.0pt}
\renewcommand{\arraystretch}{1.02}
\begin{tabular*}{\columnwidth}{@{\extracolsep{\fill}}lrrrr@{}}
\toprule
& \multicolumn{3}{c}{\ghA{VES by training stage}}
& \multicolumn{1}{c}{\ghB{Gain}} \\
\cmidrule(lr){2-4}\cmidrule(l){5-5}
\textbf{Backbone}
& \multicolumn{1}{c}{\textbf{Prompted}}
& \multicolumn{1}{c}{\textbf{Raw}}
& \multicolumn{1}{c}{\textbf{Verified}}
& \multicolumn{1}{c}{\textbf{$\Delta$V$-$R}} \\
\midrule
Qwen3-8B     & 0.594 & 0.620 & \best{0.858} & \poschg{$+0.238$} \\
Llama-3.1-8B & 0.566 & 0.496 & \best{0.649} & \poschg{$+0.153$} \\
GLM-4-9B     & 0.643 & 0.607 & \best{0.686} & \poschg{$+0.079$} \\
\bottomrule
\end{tabular*}
\caption{\textbf{Cross-backbone transfer.} Same data,
recipe, and frozen protocol; 95\% CIs: $+0.153$ $[0.109,0.197]$,
$+0.079$ $[0.036,0.120]$. Full components in
Appendix~\ref{app:additional}, Table~\ref{tab:backbone-transfer}.}
\label{tab:backbone-main}
\end{table}

The second half of the question is whether the distilled planner
can stand beside the frontier policies it learned from, so we
run each teacher as an online policy in exactly the same loop
(Table~\ref{tab:sameloop}). RefineCut-Evo ends above GPT-5.4
($+0.030$ VES, 95\% CI
$[0.001,0.062]$) and Qwen3-Max ($+0.150$, $[0.091,0.213]$),
statistically ties DeepSeek-V4-Pro ($-0.012$, $[-0.034,0.010]$),
and matches the best Converged@3 at $11.7$\,s/task locally; four
newer frontier policies land at $0.933$--$0.943$ under the same
contract (Appendix~\ref{app:additional}).

\begin{table}[t]
\centering
\scriptsize
\setlength{\tabcolsep}{2.7pt}
\renewcommand{\arraystretch}{1.02}
\begin{tabular*}{\columnwidth}{@{\extracolsep{\fill}}lrrr@{}}
\toprule
\textbf{Policy (identical loop)}
& \multicolumn{1}{c}{\textbf{VES}}
& \multicolumn{1}{c}{\textbf{Hard}}
& \multicolumn{1}{c}{\textbf{Cvg}} \\
\midrule
\rowcolor{gray!6} \multicolumn{4}{@{}l}{\textit{Frontier teachers (online)}} \\
GPT-5.4         & 0.893 & 0.76 & 0.86 \\
Qwen3-Max       & 0.773 & 0.63 & 0.72 \\
DeepSeek-V4-Pro & \best{0.936} & \best{0.89} & \best{0.95} \\
\addlinespace[1pt]
\midrule
\rowcolor{blue!5} \multicolumn{4}{@{}l}{\textit{Untrained Qwen3-8B: prompting and search}} \\
Direct prompting ($T{=}1$)        & 0.502 & 0.07 & 0.08 \\
$+$ verifier feedback ($T{\le}3$) & 0.592 & 0.15 & 0.21 \\
$R{=}4$ verifier reranking        & 0.700 & 0.32 & 0.42 \\
Visited-pool oracle               & 0.712 & 0.32 & 0.40 \\
\addlinespace[1pt]
\midrule
\rowcolor{green!8} \textbf{RefineCut-Evo (8B, local)} & 0.924 & 0.82 & \best{0.95} \\
\bottomrule
\end{tabular*}
\caption{\textbf{Same-loop external comparison.} All
policies use the identical state, ledger, RefinePatch interface,
Apply/Verify loop, stopping rule, and $T{=}3$ budget. Full tables
with latency, W/T/L, CIs, and $R{=}1$ sampling are in
Appendix~\ref{app:additional}.}
\label{tab:sameloop}
\end{table}

The lower block rules out the cheaper explanations. Verifier
feedback lifts direct prompting from $0.502$ to $0.592$; $R{=}4$
verifier reranking reaches $0.700$; and even an oracle that
picks the best state any of these runs ever visited stops at
$0.712$, still $0.212$ below Evo ($[0.170,0.252]$). Searching
over the verifier's signal is not a substitute for learning
from it.

\subsection{Robustness and Failure Analysis}
\label{sec:exp-analysis}

Aggregate numbers can hide the wrong kind of win: a gain that
leans on task overlap, is bought with heavier editing, exists
only under our own metric, or never engages with what the clips
show. The checks below rule these out in turn.

\paragraph{Canonical-clean and human-written briefs.}
The gain survives both the split and the instruction style. On
canonical-clean ($N{=}92$), where no test task shares a
canonical id with training, Evo scores $0.917$ against $0.859$
for Mixed-Pref with the variant ordering unchanged
(Table~\ref{tab:canon-clean-full}); on Common-100 each of the
three task families improves individually, the local-repair
family most of all (Table~\ref{tab:by-family}).
On Human50, $50$ free-form briefs written by seven contributors
unaffiliated with the authors (ledgers drafted by an LLM,
revised by two research assistants, and checked by two video
editors), the ranking holds and Evo stays $0.054$ ahead
($0.902$ vs.\ $0.848$; Appendix~\ref{app:human}).

\begin{figure}[t]
\centering
\includegraphics[width=.78\columnwidth]{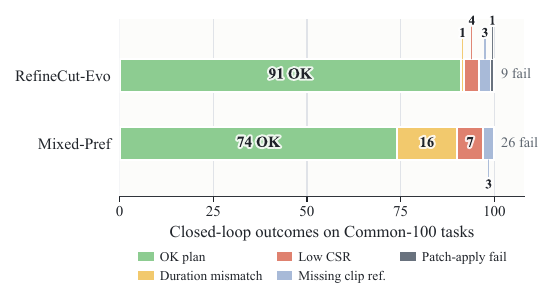}
\caption{\textbf{Failure composition.}
RefineCut-Evo increases OK plans from 74 to 91 and reduces duration
mismatches from 16 to 1 on Common-100; the small residual tail shows
all remaining closed-loop failure types.}
\label{fig:failure-mini}
\end{figure}

\paragraph{Failure composition.}
Per task, Evo mostly converts near-misses: OK plans rise from
$74$ to $91$, and duration mismatches, the dominant failure
under Mixed-Pref, fall from $16$ to $1$
(Figure~\ref{fig:failure-mini}; per-type counts in
Table~\ref{tab:failure}).

\paragraph{Plan statistics.}
The improvement is not bought with more editing. Evo uses fewer
patches per task ($1.76$ vs.\ $2.72$ for the prompted baseline),
fewer operations ($3.77$ vs.\ $5.95$), and similar output length
($67.5$ vs.\ $69.1$), and it never emits an invalid clip
reference or an over-rewrite (Table~\ref{tab:planstats}).

\paragraph{LLM-judge agreement.}
Does an evaluator outside our loop see the same ordering? A
blind API judge points the same way as the verifier on
$68$--$70\%$ of tasks with a nonzero VES margin, and on every
task the verifier marks as a degradation.
Its many ties sit almost entirely on the $73/100$ tasks whose
terminal plans also tie under the verifier
(Appendix~\ref{app:judge-validity}).

\paragraph{Semantic dependence.}
If the planner ignored the captions, editing them should not
matter. It does not ignore them: removing all clip semantics
costs $0.074$ VES, and shuffling captions across clips costs
$0.200$, with required-clip recall collapsing from $0.98$ to
$0.46$. Deleting only the short caption, whose content overlaps
the structured fields, changes nothing measurable (full deltas
and CIs in Appendix~\ref{app:additional}).

\subsection{Blind Rendered-Preview Preference}
\label{sec:exp-human}

Do the planning gains survive rendering? Three annotators per
pair compared $150$ blind, left--right randomized A/B previews
for each model pair (protocol in Appendix~\ref{app:human}). Evo
beats Mixed-Pref in $100$ of the $150$ pairs, ties $34$, and
loses $16$ (preference $0.780$, $\kappa{=}0.620$); the sanity
pair Mixed-Pref vs.\ Prompted, whose gap is large, reaches
$0.887$. Preference here scores the planner and renderer
together, and it agrees with the VES ordering.

\subsection{Summary of Findings}
\label{sec:exp-summary}

Verifier-replayed supervision provides the largest single
improvement over raw imitation and transfers across Qwen3-8B,
Llama-3.1-8B, and GLM-4-9B. In the identical closed
loop, RefineCut-Evo outperforms two of its three teachers, ties
the third, and stays well above prompting and inference-time
search on the same backbone. Human briefs, rendered previews,
plan statistics, and sensitivity and fallback analyses
corroborate these results beyond one backbone, variant, or
scoring configuration.

\section{Conclusion}
\label{sec:conclusion}

We treated video editing as a planning problem whose outputs can
be checked, and built RefineCut around that property. An
explicit constraint ledger turns each brief into
machine-checkable requirements, a typed timeline with
RefinePatch operations makes edits executable, and a
deterministic verifier closes the loop. The same verifier then
does the training work: replaying every candidate branch turns
noisy multi-teacher traces into verified supervision, and
RefineCut-Evo continues improving the planner on its own
verified repairs with high-margin DPO. On RefineCut-Bench this
recipe lifts an $8$B planner from $0.620$ to $0.924$ VES, the
verified-over-raw gain carries over to Llama-3.1-8B and
GLM-4-9B, and the final planner stands beside the frontier
teachers it distilled from, inside the same closed loop. The
broader lesson: when a creative task admits an executable
specification, a deterministic verifier can stand in for human
labels or a learned judge as the primary training signal, and a
compact open model can then be trained to hold its own against
much larger prompted systems.
\section*{Limitations}

\paragraph{Planning-layer scope.}
RefineCut is deliberately scoped to executable edit planning over
a typed timeline. The verifier checks structural requirements,
from schema validity to ledger satisfaction and duration control;
it cannot judge whether a cut is tasteful or a story lands, which
only the rendered-preview study touches. Because the verifier
also supplies training-time replay and closed-loop feedback, VES
is an in-protocol measure rather than an independent absolute
quality score.

\paragraph{Upstream perception.}
The planner reads upstream captions, motion metadata, and music
metadata rather than raw pixels, so caption or beat-tracking errors
can bound edit-plan quality. Storyboard previews also mix planner
effects with the rendering pipeline. Controlled removal and shuffling
of planner-visible semantic fields show that the planner uses textual
clip semantics, but do not establish robustness to errors from a real
captioner, robustness across captioners, or raw-pixel perception;
these remain future work.

\paragraph{Generalization and future work.}
The current evidence covers three model families in the compact
8B--9B regime, three editing-task families, and one primary asset
pool; the preference stages are trained only on Qwen3-8B, and
absolute transfer scores remain below it. RefineCut-Evo is an
offline DPO stage rather than a full EvoLM reproduction or online
RL. Task briefs and constraint ledgers in RefineCut-Bench are
LLM-generated under controlled templates, which keeps
verifier-based evaluation reproducible; Human50 tests transfer to
free-form human-written briefs, while larger-scale real-user
specifications, additional asset domains, model scales,
longer-form editing, and final-video evaluation remain future
work.

\section*{Ethical Considerations}

RefineCut studies planning-level editing rather than unrestricted
video generation, but executable edit plans can connect to rendering
or generation tools and could be misused to create misleading edits,
omit context, or insert synthetic content without disclosure. Our
benchmark uses controlled tasks, explicit ledgers, and clip-level
metadata; deployment on real media should respect copyright,
consent, provenance, and disclosure requirements. Clips come from
public research-licensed sources and are not annotated with person
identities; captions are short scene/action descriptions and do
not contain personal names. The planner reads only captions and
metadata, not raw pixels. The verifier checks structural
correctness, not factual accuracy, fairness, or social
appropriateness, so we view RefineCut as a research framework for
verifier-guided planner training and encourage future work on
provenance tracking, misuse detection, and oversight.

\section*{Data Availability}
\phantomsection\label{sec:availability}

The code is available at \url{\codeurl} and RefineCut-Bench at
\url{\artifacturl}. The benchmark release contains the $3{,}578$
canonical tasks with briefs, constraint ledgers, clip-pool metadata
and captions, music metadata with beat tracks, all splits including
Common-100, dev100, and the canonical-clean subset, the canonicalized
multi-teacher trajectories with per-branch verifier replay scores,
and the JSON schemas for the constraint ledger, edit plan,
RefinePatch, timeline IR, and verifier output. The code release
contains the deterministic verifier, RefinePatch canonicalization
and the Apply/Verify loop, the metric implementation, the
PatchPlanner evaluation prompt, and the closed-loop evaluation
harness used for all reported numbers. Raw video clips and music are
referenced by their public research-licensed source identifiers
rather than redistributed.

\section*{Declaration of Generative AI Usage}

During preparation, the authors used AI assistants for language
polishing, LaTeX formatting, consistency checking, and coding
assistance. AI tools did not generate experimental results or
automatically validate references. All claims, analyses, citations,
and final decisions were manually verified and approved by the
authors, who bear full responsibility for the manuscript.

\section*{Acknowledgments}

This work is supported in part by the Guangdong Basic and Applied
Basic Research Foundation under Grant No.~2025A1515012968, in part
by the Shenzhen Science and Technology Program under Grant
No.~JCYJ20240813113502004, in part by the National Natural Science
Foundation of China under Grant No.~62001412, in part by Shenzhen
Stability Science Program 2023, in part by the Guangdong Provincial
Key Laboratory of Future Networks of Intelligence (Grant
No.~2022B1212010001), and in part by the Shenzhen Key Lab of Crowd
Intelligence Empowered Low-Carbon Energy Network (Grant
No.~ZDSYS20220606100601002).

\bibliography{custom2}

\clearpage
\newpage

\appendix


\section{Related Work}
\label{app:related}

\subsection{AI-Assisted Video Editing and Generation}
A first line of work focuses on creating or modifying visual
content. Text-to-video and image-to-video models generate clips
de novo \citep{zhuang2024vlogger,long2024videostudio,yang2025cogvideox,kong2024hunyuanvideo,wan2025wan,polyak2024moviegen},
and instruction-based editing models change appearance or style
within a fixed shot or short sequence
\citep{cheng2024insv2v,ku2024anyv2v,yoon-etal-2025-raccoon}.
Trailer, montage, and mashup systems
\citep{li2026direct,lin2026glance,zhu2025weakly}
assemble cuts from a footage library according to a script or
theme, and UniVA \citep{liang2025univa} packages generation,
editing, and segmentation under one tool-calling loop. These systems can produce or rearrange content. However, they do
not train a reusable open-weight planner: a planner that learns
editing decisions over a real clip pool from a constraint ledger
and an editable timeline.

\subsection{Workflow Decomposition for Multimodal Creation}
Hierarchical multi-agent systems break complex creation into
staged professional roles. DIRECT \citep{li2026direct} decomposes
mashup creation into Screenwriter, Director, and Editor agents
under a hierarchical multimodal coherency objective.
LVAS-Agent \citep{zhang2025lvasagent} mirrors a long-video
dubbing studio with Storyboarder, Scriptwriter, Designer, and
Generator. StoryAgent \citep{hu2024storyagent} takes a similar
role-based view of storytelling video production. These systems
share two properties: the planning logic is built from prompted
frontier backbones, and the editing decision flow is encoded in
the prompt scaffolding rather than learned. RefineCut keeps the
workflow-decomposition view but inverts the operating mode. We
collect trajectories from such workflows, replay them through a
deterministic verifier, and train an open-weight planner that
internalizes the editing decision flow.

\subsection{Tool-Executable Planners from Trajectories}
A complementary line of work studies agents that reason, act, or
revise through tool-like trajectories. Prompt-time and
self-correction methods such as ReAct~\citep{yao2023react},
Self-Refine~\citep{madaan2023selfrefine}, and
Reflexion~\citep{shinn2023reflexion} demonstrate iterative
reasoning or feedback use without necessarily updating model
weights. Training-oriented systems instead use interaction or
tool-use trajectories to improve agent behavior:
AgentTuning~\citep{zeng-etal-2024-agenttuning} fine-tunes on
tool-use traces, AgentBank~\citep{song2024agentbank} scales
trajectory fine-tuning across diverse agent skills, and
AgentGym~\citep{xi2025agentgym} evolves agents across
environments from collected trajectories. Trajectory feedback
can also be turned into a preference signal:
ETO~\citep{song2024eto} pairs successful and failed trajectories
for Direct Preference Optimization (DPO)~\citep{rafailov2023dpo},
IPR~\citep{xiong2024watch} estimates step-level rewards by
Monte-Carlo sampling along expert trajectories, and
negative-aware training~\citep{wang2025nat} incorporates failed
trajectories during fine-tuning. RefineCut belongs to this
broader executable-agent line, but differs by using
deterministic replay of typed video-editing
patches~\citep{rfc6902} to construct both supervised fine-tuning
(SFT) targets and preference pairs.

\subsection{Rubric-Guided and Verifier-Guided Self-Improvement}
Recent work explores rubric-structured self-improvement of
language models. EvoLM \citep{li2026evolmselfevolvinglanguagemodels}
uses a co-trained rubric model to score self-generated candidates
and drive student improvement. Rubric-Grounded RL
\citep{bhattarai2026rubricgroundedrlstructuredjudge} applies
multi-criterion weighted rewards, and UCPO-style diversity
penalties \citep{lochab2026ucpo} target collapse in
multiple-correct settings. RefineCut adapts this idea to
executable video-editing planning. The student samples candidate
repairs at training states, each candidate is scored by the
deterministic verifier together with a task-specific editing
rubric, and high-margin pairs train the planner with DPO: EvoLM's
co-trained rubric model is replaced by a deterministic verifier
plus a fixed task rubric, and UCPO-lite appears only as a
negative diagnostic in the ablation.

\paragraph{Preference-optimization objectives.}
RefineCut uses DPO because verifier replay and Evo candidate
scoring naturally yield paired chosen/rejected repairs. Recent
alternatives such as IPO~\citep{azar2024ipo},
KTO~\citep{ethayarajh2024kto}, ORPO~\citep{hong2024orpo}, and
SimPO~\citep{meng2024simpo} modify the preference objective,
reduce reference-model dependence, or use
unary/desirable--undesirable signals. These objectives are
orthogonal to our central contribution: deterministic verifier
replay creates the supervision source. We therefore use DPO as a
standard paired-preference optimizer and leave objective swaps to
future work.

\section{RefineCut-Bench Construction Details}
\label{app:bench-details}

This appendix lists the construction steps required to reproduce
RefineCut-Bench: asset collection, captioning, task
generation, ledger design, clip-pool sampling, teacher trajectory
collection, and the canonical-id split. The main benchmark overview
appears in Figure~\ref{fig:bench}.

\begin{table}[t]
\centering
\scriptsize
\renewcommand{\arraystretch}{1.05}
\begin{tabular*}{\columnwidth}{@{\extracolsep{\fill}}lr@{}}
\toprule
\textbf{Item} & \textbf{Value} \\
\midrule
Raw video clips                   & 7{,}971 \\
Extracted caption frames          & 23{,}913 \\
Music tracks                      & 499 \\
\midrule
Raw task records                  & 3{,}960 \\
Canonical unique tasks            & 3{,}578 \\
Train / Dev / Test (records)      & 2{,}773 / 596 / 591 \\
Test canonical-clean              & 518 / 591 \\
Removed by canonical overlap      & 73 ($12.35\%$) \\
Common-100 $\cap$ canon.-clean    & 92 \\
Task-level record overlap         & 0 / 0 / 0 \\
\midrule
Raw teacher rollouts              & 2{,}000 / 1{,}962 / 1{,}962 \\
Replayed teacher rollouts         & 2{,}000 / 1{,}946 / 1{,}959 \\
Raw canonical triple intersection          & 381 \\
Replayed canonical triple intersection     & 375 \\
Canonical teacher union                    & 3{,}578 \\
\bottomrule
\end{tabular*}
\caption{Dataset, splits, and teacher coverage. Teacher counts
are GPT-5.4 / Qwen3-Max / DeepSeek-V4-Pro.}
\label{tab:split}
\end{table}

\subsection{Asset Pool and Caption Sources}
\label{app:bench-assets}

The asset pool contains $7{,}971$ captioned clips drawn from five
public sources (Table~\ref{tab:caption-sources}). Each row in the
clip caption table carries the fields \texttt{clip\_id},
\texttt{source}, \texttt{duration}, \texttt{subject},
\texttt{action}, \texttt{scene}, \texttt{camera},
\texttt{scene\_category}, \texttt{motion\_intensity},
\texttt{caption\_short}, \texttt{model\_used}, and \texttt{ts}.
The \texttt{model\_used} field records which caption model emitted
each row, so captioner provenance is tracked per clip rather than
assumed globally; the supplementary artifact records the exact
\texttt{model\_used} distribution.

\begin{table}[ht]
\centering
\footnotesize
\renewcommand{\arraystretch}{1.05}
\begin{tabular*}{\columnwidth}{@{\extracolsep{\fill}}lr@{}}
\toprule
\textbf{Source} & \textbf{Captioned clips} \\
\midrule
panda70m\_sample      & 3{,}990 \\
pexels                & 3{,}219 \\
pixabay               & 255 \\
openvid               & 100 \\
long\_video\_segment  & 407 \\
\midrule
Total                 & 7{,}971 \\
\bottomrule
\end{tabular*}
\caption{Captioned clip counts per source. The total matches the
asset count reported in Section~\ref{sec:bench}.}
\label{tab:caption-sources}
\end{table}

\subsection{Frame Sampling and Captioning}
\label{app:bench-frames}

We use an adaptive frame sampler (``v2'') that estimates motion
intensity from optical-flow magnitude on a coarse temporal grid
and then samples $3$, $5$, or $7$ frames per clip for
low / medium / high motion. The motion bucket is also stored in
the \texttt{motion\_intensity} field of the caption row, so that
downstream task generation can use it as a metadata signal.

The captioning module is an upstream component: the planner does
not read pixels and only sees the caption fields. The captioner
varies across rows; each row's exact configuration is recorded in
\texttt{model\_used} and released with the dataset.

\subsection{Task Generation}
\label{app:bench-taskgen}

Tasks are generated by an LLM conditioned on a sampled clip pool,
the family and subtype slot in Table~\ref{tab:subtypes}, and a
family-conditioned ledger template. The generated record separates
task identity, the natural-language brief, planner inputs, and
subtype-specific structure.

\begin{center}
\begin{minipage}{\columnwidth}
\centering
\scriptsize
\setlength{\tabcolsep}{2.5pt}
\renewcommand{\arraystretch}{1.02}
\begin{tabularx}{\columnwidth}{@{}l>{\raggedright\arraybackslash}X@{}}
\toprule
\textbf{Record group} & \textbf{Fields} \\
\midrule
Identity &
\texttt{task\_id}, \texttt{task\_type}, \texttt{task\_subtype} \\
Brief &
\texttt{brief}, \texttt{target\_duration} \\
Planner inputs &
\texttt{clip\_pool}, \texttt{constraint\_ledger} \\
Structure &
\texttt{structure}, \texttt{subtype\_constraints} \\
\bottomrule
\end{tabularx}
\vspace{1pt}
\par\raggedright\scriptsize\emph{Task-record schema.}
The appendix names implementation fields here; the surrounding text
uses plain-language descriptions.
\end{minipage}
\end{center}

\begin{table}[ht]
\centering
\scriptsize
\renewcommand{\arraystretch}{1.05}
\begin{tabular*}{\columnwidth}{@{\extracolsep{\fill}}ll@{}}
\toprule
\textbf{Family} & \textbf{Subtypes} \\
\midrule
A (multi-clip   & themed\_montage, pacing\_progression, \\
composition)    & motion\_dynamics\_montage, scene\_traversal, \\
                & character\_focus, abstract\_rhythm \\
\midrule
B (local        & b\_roll\_insert, clip\_swap, cut\_extend, \\
edit/repair)    & transition\_repair \\
\midrule
C (generative   & generated\_opener, generated\_bridge, \\
composition)    & generated\_text\_overlay, \\
                & generated\_replacement, \\
                & generated\_full\_assembly \\
\bottomrule
\end{tabular*}
\caption{The $15$ task subtypes grouped by family.}
\label{tab:subtypes}
\end{table}

\subsection{Constraint Ledger Design}
\label{app:bench-ledger}

The constraint ledger is the explicit specification of what a
successful cut has to satisfy. To keep the appendix readable, we
group the $14$ fine-grained ledger types by the editing requirement
they express.

\begin{center}
\begin{minipage}{\columnwidth}
\centering
\scriptsize
\setlength{\tabcolsep}{2.5pt}
\renewcommand{\arraystretch}{1.02}
\begin{tabularx}{\columnwidth}{@{}l>{\raggedright\arraybackslash}X@{}}
\toprule
\textbf{Requirement} & \textbf{Ledger types} \\
\midrule
Duration &
\texttt{target\_duration}, \texttt{duration\_tolerance} \\
Transition &
\texttt{transition\_style} \\
Music sync &
\texttt{music\_sync\_bpm}, \texttt{music\_sync\_beat} \\
Clip inclusion &
\begin{tabular}[t]{@{}l@{}}
\texttt{must\_keep\_clip}, \texttt{must\_open\_with}\\
\texttt{tag\_inclusion}
\end{tabular} \\
Clip exclusion &
\begin{tabular}[t]{@{}l@{}}
\texttt{must\_exclude\_clip}, \texttt{must\_close\_with}\\
\texttt{tag\_exclusion}
\end{tabular} \\
Repeat limit &
\begin{tabular}[t]{@{}l@{}}
\texttt{no\_repeat\_within\_seconds}\\
\texttt{max\_repeats\_per\_clip}
\end{tabular} \\
Pacing &
\texttt{pacing} \\
\bottomrule
\end{tabularx}
\vspace{1pt}
\par\raggedright\scriptsize\emph{Ledger type groups.}
Fine-grained types are shown as implementation field names; the
main text uses the broader requirement names.
\end{minipage}
\end{center}

Every ledger ships with at least four items, and every ledger
includes a \texttt{target\_duration} item. The seven field-level
constraint families mentioned in Section~\ref{sec:task-io}
(duration, transition, music synchronization, clip inclusion, clip
exclusion, repeat limit, and pacing) group these field-level
checks; generated-placeholder requirements are checked separately
as task-structure constraints.
Per-type counts on the $3{,}578$ canonical task universe are shown
in Figure~\ref{fig:bench}~(C); the full numeric table is released
with the dataset.

\subsection{Clip-Pool Sampling}
\label{app:bench-pool}

The \texttt{clip\_pool} of a task is sampled from the captioned
asset pool under the following rules.

(i)~The planner may only choose clips from the given
\texttt{clip\_pool}; references to a \texttt{clip\_id} outside the
pool are flagged as invalid by the verifier.

(ii)~When the candidate pool ($\textit{given\_pool}$) contains at
least $30$ clips, the sampled \texttt{clip\_pool} size is between
$30$ and $60$; otherwise the entire candidate pool is used.

(iii)~The ledger is generated jointly with the
\texttt{clip\_pool} so that every \texttt{must\_keep\_clip} and
\texttt{must\_exclude\_clip} item refers to a clip that exists in
the pool.

\subsection{Teacher Trajectory Generation}
\label{app:bench-traj}

For the teacher-covered subset of canonical tasks we collect a
teacher trajectory from each available frontier API teacher
(GPT-5.4, Qwen3-Max, and DeepSeek-V4-Pro); coverage is partial,
and the per-teacher replayed counts and canonical triple
intersection are reported in Table~\ref{tab:split}.
At every refine step a teacher must emit exactly four
candidate branches; each branch is a \texttt{RefinePatch} that
carries: a list of \texttt{operations} (RFC~6902 add / remove /
replace), a \texttt{rationale\_against\_ledger} naming the ledger
items it tries to repair, a \texttt{repair\_operator} label from
the fixed vocabulary, a free-text \texttt{thought}, and a numeric
\texttt{confidence}.

\begin{table}[ht]
\centering
\footnotesize
\renewcommand{\arraystretch}{1.05}
\begin{tabular*}{\columnwidth}{@{\extracolsep{\fill}}ll@{}}
\toprule
\textbf{\texttt{repair\_operator}} & \textbf{Intent} \\
\midrule
\texttt{modify\_prompt}      & rewrite the generative prompt \\
\texttt{change\_tool}        & switch the downstream tool slot \\
\texttt{adjust\_parameter}   & tweak a numeric or enum field \\
\texttt{reselect\_clip}      & replace a clip in the sequence \\
\texttt{retry\_with\_fallback} & redo with a safer default \\
\texttt{abort\_and\_skip}    & drop the step \\
\bottomrule
\end{tabular*}
\caption{The six \texttt{repair\_operator} values.}
\label{tab:repair-ops}
\end{table}

\subsection{Splits and Canonical IDs}
\label{app:bench-splits}

Raw task records are deduplicated to canonical task ids by
collapsing minor surface variants (re-sampled \texttt{clip\_pool}s
of the same underlying task) into a single canonical id. We then
split at the record level into train / dev / test of
$2{,}773 / 596 / 591$. Of the $591$ test records, $518$ carry a
canonical id that does not appear in the train set
(\emph{canonical-clean}); the remaining $73$ records overlap by
canonical id with train. The intersection of canonical-clean with
the Common-100 evaluation subset is $N=92$ records and is used
for the robustness check in Section~\ref{sec:exp-analysis}.

\subsection{Licensing and Intended Use}
\label{app:bench-license}

All clip sources in Table~\ref{tab:caption-sources} are publicly
available and permit research use; the music tracks and a subset
of in-house clips are private assets used for research only. We
use every source within its stated terms and solely for
non-commercial research. RefineCut-Bench (task records, ledgers,
captioned-clip and music metadata, multi-teacher trajectories)
and the training/inference code will be released under
CC~BY-NC~4.0 upon acceptance. Derived artifacts inherit the
research-only restriction of their underlying sources and should
not be used outside research contexts.

\section{Schemas and Interfaces}
\label{app:schemas}

This appendix gives compact schemas for the five interfaces that
appear throughout the paper. Full JSON Schemas and longer worked
examples are in the supplementary release.

\begin{schemabox}[sbblue]{RefinePatch -- a planner step}
\begin{lstlisting}[language=jsonbare]
{
  "operations": [
    {"op": "replace",
     "path": "/args/selected_sequence",
     "value": ["C003","C001","C007","C012"]},
    {"op": "replace",
     "path": "/args/segment_durations",
     "value": [2.0, 3.5, 4.0, 3.5]}
  ],
  "rationale_against_ledger":
    ["target_duration#0", "must_keep_clip#2"],
  "repair_operator": "reselect_clip",
  "affected_clips": ["C001","C007","C012"],
  "thought": "Swap C005 -> C007 to keep duration.",
  "confidence": 0.71
}
\end{lstlisting}
\end{schemabox}

\begin{schemabox}[sbgreen]{EditPlan / Timeline IR -- closed-loop terminal artifact}
\begin{lstlisting}[language=jsonbare]
{
  "selected_sequence": ["C003","C001","C007","C012"],
  "segment_durations": [2.0, 3.5, 4.0, 3.5],
  "transitions": ["cut","cross_dissolve","cut"],
  "music_sync": {"bpm": 96, "beat_offset_ms": 120},
  "placeholders": [],
  "renderer_hints": {"target_fps": 30}
}
\end{lstlisting}
\end{schemabox}

\begin{schemabox}[sborange]{ConstraintLedger entry -- one row}
\begin{lstlisting}[language=jsonbare]
{
  "item_id": "target_duration#0",
  "type": "target_duration",
  "spec": {"value_sec": 13.0, "tolerance_sec": 2.0},
  "satisfied": false,
  "evidence": {"current_sec": 15.4}
}
\end{lstlisting}
\end{schemabox}

\begin{schemabox}[sbpurple]{VerifierOutput -- one verifier call}
\begin{lstlisting}[language=jsonbare]
{
  "schema_ok": true, "patch_apply_ok": true,
  "timeline_valid": true,
  "delta_csr": 0.20,
  "targeted_repair": true,
  "no_regression": true,
  "req_clip_recall": 1.0,
  "locality": 1.0,
  "score": 0.72
}
\end{lstlisting}
\end{schemabox}

\begin{schemabox}[sbteal]{RubricScore -- Evo rubric weighted aggregate ER1--ER7 (illustrative per-task instance)}
\begin{lstlisting}[language=jsonbare]
{
  "ER1": 0.50, "ER2": 1.00, "ER3": 0.75,
  "ER4": 0.50, "ER5": 1.00, "ER6": 0.50, "ER7": 1.00,
  "weights":
    {"ER1":0.10,"ER2":0.25,"ER3":0.20,
     "ER4":0.10,"ER5":0.15,"ER6":0.05,"ER7":0.15},
  "total": 0.825,
  "fallback_used": ["ER1", "ER3_judge"]
}
\end{lstlisting}
\end{schemabox}

\section{Metric Definitions}
\label{app:metrics}

\paragraph{Execution Gate.}
\textsc{SCR} (schema-conformance rate): output parses against the
RefinePatch JSON schema. \textsc{PASR} (patch-apply success
rate): canonicalized patch applies cleanly to $s_t$.
\textsc{TimelineValidity}: the resulting state serializes to a
Timeline IR that the renderer parser accepts.

\paragraph{Planning Quality.}
\textsc{FinalCSR} (FCSR; terminal constraint-satisfaction rate):
terminal mean per-entry pass rate over the ledger.
\textsc{HardPass} (Hard): $\mathbb{1}[\textsc{FinalCSR}{=}1]$.
\textsc{TargetedRepair}: fraction of ledger items named in the
patch's \texttt{rationale\_against\_ledger} that are satisfied
after \texttt{Apply}. \textsc{FinalReqClipRecall} (ReqCR): recall
over the required clip ids on the terminal sequence (denominator
rules in Appendix~\ref{app:verifier}). \textsc{ValidClipPrecision}:
fraction of clip references in the terminal sequence that lie
inside the pool. \textsc{DurationPass@2s} (Dur):
$\mathbb{1}[|\textsc{Dur}_\text{final}-\textsc{Dur}_\text{target}|\le 2]$.
\textsc{NoRegressionAllSteps}: no previously satisfied ledger
item becomes unsatisfied at any step.
\textsc{Converged@3} (Cvg): $\mathbb{1}[\textsc{FinalCSR} \ge 0.8]$
within $T{=}3$ closed-loop steps (operational rule in
Appendix~\ref{app:verifier}).

\paragraph{VES summary.}
\textsc{VES} is a fixed weighted sum:
\begin{equation*}
\setlength{\jot}{1pt}
\begin{aligned}
\mathrm{VES}={}&0.30\,\mathrm{FinalCSR}\\
&+0.15\,\mathrm{HardPass}
+0.15\,\mathrm{PASR}\\
&+0.15\,\mathrm{FinalReqClipRecall}\\
&+0.10\,\mathrm{DurationPass}\\
&+0.10\,\mathrm{TimelineValidity}\\
&+0.05\,\mathrm{NoRegression}.
\end{aligned}
\end{equation*}

\paragraph{Branch-score signals.}
The replay score of a single candidate branch combines six
signals. \textsc{$\Delta$CSR} is the increase in the per-entry
ledger pass rate between the pre-patch state $s_t$ and the
post-patch state $s_{t+1}$. \textsc{TargetedRepair},
\textsc{ReqClipRecall}, \textsc{PASR}, and \textsc{NoRegression}
are the per-branch counterparts of the planning-quality metrics
above, evaluated for one applied branch rather than the terminal
state. \textsc{Locality} is the fraction of patch operations that
edit only fields tied to the ledger items named in
\texttt{rationale\_against\_ledger}; it penalizes patches that
modify state unrelated to the items being repaired.

\paragraph{Branch score weights.}
The replay branch score uses the six signals named in the
method section:
\begin{equation}
\begin{aligned}
V(b;s_t,L_t)={}&
w_1\Delta\mathrm{CSR}\\
&+w_2\mathrm{TargetedRepair}\\
&+w_3\mathrm{ReqClipRecall}\\
&+w_4\mathrm{PASR}\\
&+w_5\mathrm{NoRegression}\\
&+w_6\mathrm{Locality}.
\end{aligned}
\label{eq:branch-score}
\end{equation}
It uses fixed weights
$(w_1, w_2, w_3, w_4, w_5, w_6)$ $=$
$(0.35, 0.20, 0.20, 0.10, 0.10, 0.05)$
applied in order to
$\Delta\mathrm{CSR}$, \textsc{TargetedRepair},
\textsc{ReqClipRecall}, \textsc{PASR}, \textsc{NoRegression},
and \textsc{Locality}.

\paragraph{Sensitivity analyses.}
Table~\ref{tab:sensitivity} summarizes sensitivity to the loop
budget $T$, the candidate count $K$, the mix weight $\lambda$,
and the VES weights. The $T$ and $K$ analyses are post-hoc over
the archived trajectories and candidate pools, not new inference
or training.

\begin{center}
\begin{minipage}{\columnwidth}
\centering
\scriptsize
\setlength{\tabcolsep}{3pt}
\renewcommand{\arraystretch}{1.05}
\begin{tabular*}{\columnwidth}{@{\extracolsep{\fill}}lll@{}}
\toprule
\textbf{Dimension} & \textbf{Setting} & \textbf{Result} \\
\midrule
Loop budget $T$ & $1/2/3$ (truncation) &
\makecell[l]{Evo VES $0.677/0.908/0.924$;\\first at every prefix} \\
\addlinespace[2pt]
Mix weight $\lambda$ & $0.5/0.8/1.0$ &
\makecell[l]{Kendall $\tau$-b $\ge 0.973$;\\$0/779$ pair flips} \\
\addlinespace[2pt]
Candidates $K$ & $2/3$ from $K{=}4$ pool &
\makecell[l]{top-1 retention $0.500/0.750$\\(sampling coverage)} \\
\addlinespace[2pt]
VES weights & \makecell[l]{$14$ weights $\pm50\%$;\\$20{,}000$
reweightings} &
\makecell[l]{Evo first, above Mixed-Pref\\in $100\%$ of settings} \\
\bottomrule
\end{tabular*}
\captionsetup{hypcap=false}
\captionof{table}{\textbf{Sensitivity of the main ranking.} The
$T$ and $K$ rows are post-hoc analyses over archived trajectories
and candidate pools.}
\label{tab:sensitivity}
\end{minipage}
\end{center}

\section{Training Details}
\label{app:training}

\paragraph{Backbone and LoRA.}
All planners in the primary progression initialize from
Qwen3-8B-Instruct with LoRA \citep{hu2022lora}: rank $r{=}32$,
$\alpha{=}64$, applied to the q/k/v/o and gate/up/down
projections. The cross-backbone transfer runs
(Section~\ref{sec:exp-rq5}) reuse the identical adapter
configuration on Llama-3.1-8B-Instruct and GLM-4-9B. Adapter
checksums are released with the training artifacts.

\paragraph{Raw SFT.}
Examples $9{,}690$ (teacher first-choice branches without
verifier filtering). Max steps $800$; other hyperparameters
follow Verified SFT.

\paragraph{Verified SFT.}
Examples $3{,}317$ (one verifier-best target per replayed step
that passes the verified-SFT filter; Section~\ref{sec:method-distillation}).
Max steps $1{,}200$. Learning rate $5\mathrm{e}{-}5$.
Per-device batch $2$ with gradient accumulation $8$. Loss is
standard next-token cross-entropy on the canonicalized patch.

\paragraph{Traj-Pref DPO.}
Examples $357$ trajectory-level pairs. Max steps $400$; other
hyperparameters follow Mixed-Pref. Initialized from Verified.

\paragraph{Mixed-Pref DPO.}
Examples $2{,}380$ chosen / rejected pairs, mixing step-level and
trajectory-level signals. Max steps $800$. Learning rate
$2\mathrm{e}{-}6$. $\beta = 0.1$. Initialized from
Verified.

\paragraph{RefineCut-Evo DPO.}
Candidate generation: $K=4$ candidates per training state, on
$1{,}500$ training states. Pair construction: $779$
rubric-margin pairs oversampled to $1{,}651$ training rows
(Appendix~\ref{app:rubric}). Max steps $600$ with an intermediate
checkpoint at step $300$. Learning rate $1\mathrm{e}{-}6$.
$\beta = 0.05$. Initialized from Mixed-Pref. Step $300$
is selected on dev100 VES
(Table~\ref{tab:dev-selection}).

\paragraph{Ablation variants.}
Verifier-only, No-margin, and
UCPO-lite share the same backbone, candidate pool, and
DPO recipe as Rubric-margin; they differ only in the
pair-scoring function and the margin filter
(Table~\ref{tab:ablation}). Hyperparameters and elapsed wall-clock
times are matched across variants.

\paragraph{Compute budget and cost.}
All training ran on a single NVIDIA A100 GPU. The core pipeline
consumed approximately $14.5$ GPU-hours: Verified SFT ($1{,}200$
steps, ${\sim}7.5$~h), Mixed-Pref preference fine-tuning ($800$
steps, ${\sim}2.1$~h), and the RefineCut-Evo Rubric-margin DPO
run ($600$ steps, ${\sim}4.9$~h; the step-$300$ checkpoint is
selected by dev VES). Including the Raw SFT ablation, the total
is $23.5$ GPU-hours; the three step-$300$ DPO ablations
(Verifier-only, No-margin, UCPO-lite) each add ${\sim}2.5$
matched wall-clock hours, and generating the $6{,}000$ Evo
candidates took $3.98$~h locally with no API calls. Multi-teacher
trajectory collection used $13{,}804$ frontier API calls to
GPT-5.4, Qwen3-Max, and DeepSeek-V4-Pro
(Section~\ref{sec:bench-traj}), archiving $39.2$M output
characters (${\approx}10$--$11$M output tokens); input volume,
estimated from per-call payloads (mean $11.1$k characters), is
${\approx}46$--$92$M tokens, and the total spend was
approximately CNY $2{,}000$--$3{,}000$ (${\approx}\$280$--$420$).
Call counts and output volume are exact archival values; input
tokens and monetary cost are estimates from payload sizes and
author accounting records. At inference, RefineCut-Evo runs
locally at $11.7$\,s/task, while teacher and frontier APIs range
$6.7$--$36.9$\,s/task; per-task token usage and latency are
released with the artifacts. No model training was performed
through the APIs.

\paragraph{Decoding.}
All closed-loop evaluations use $R{=}1$ greedy decoding at
generation time (a single sample per step). $R$ is the
inference-time sampling/reranking pool size and is distinct from
the training-time student-candidate count $K{=}4$
(Section~\ref{sec:method-evo}). $R{=}4$ verifier reranking over
the untrained backbone is evaluated as a search control under the
frozen protocol (Table~\ref{tab:sameloop};
Appendix~\ref{app:additional}, Table~\ref{tab:ladder}).

\section{RefineCut-Evo Details}
\label{app:rubric}

\paragraph{Verifier replay procedure.}
Algorithm~\ref{alg:replay} formalizes the verifier replay and
branch-arbitration procedure used by the Stage~1 distillation
(Section~\ref{sec:method-replay}).

\begin{algorithm}[ht]
\small
\caption{Verifier replay and branch arbitration.}
\label{alg:replay}
\begin{algorithmic}[1]
\Require task $x$, initial state $s_0$, teacher trajectory
$\tau = (\textit{step}_1, \dots, \textit{step}_T)$
\Ensure replayed trajectory with per-branch scores
\State $s \leftarrow s_0$;\quad $L \leftarrow \textsc{InitLedger}(x, s_0)$
\For{$t = 1, \dots, T$}
  \State $R_t \leftarrow \emptyset$
  \For{each candidate branch $b_t^k$ in $\textit{step}_t$}
    \State $p_t^k \leftarrow \textsc{Canonicalize}(b_t^k)$
    \If{\textbf{not} $\textsc{Valid}(p_t^k, s)$}
      \State $r_t^k \leftarrow \textsc{Invalid}$;\, \textbf{continue}
    \EndIf
    \State $s' \leftarrow \textsc{Apply}(s, p_t^k)$
    \State $L' \leftarrow \textsc{Recompute}(x, s')$
    \State $r_t^k \leftarrow \textsc{Score}(s, s', L, L')$ \Comment{Eq.~\ref{eq:branch-score}}
    \State $R_t \leftarrow R_t \cup \{(p_t^k, r_t^k)\}$
  \EndFor
  \State $b_t^* \leftarrow \arg\max_k r_t^k$
  \State $s \leftarrow \textsc{Apply}(s, p_t^*)$
  \State $L \leftarrow \textsc{Recompute}(x, s)$
\EndFor
\State \Return replayed trajectory with $\{R_t, b_t^*\}_{t=1}^T$
\end{algorithmic}
\end{algorithm}

\paragraph{Rubric construction.}
For each canonical task we construct a task-specific editing
rubric with seven criteria ER1--ER7
(Section~\ref{sec:method-evo}). The criteria and their
average weights across the train set are listed in
Table~\ref{tab:rubric-weights}. Weights are computed per
task-family from a template that puts mass on the criteria most
relevant to the family.

\begin{table}[ht]
\centering
\footnotesize
\renewcommand{\arraystretch}{1.05}
\begin{tabular*}{\columnwidth}{@{\extracolsep{\fill}}llc@{}}
\toprule
\textbf{ID} & \textbf{Criterion} & \textbf{Avg.\ weight} \\
\midrule
ER1 & Intent and Story Alignment        & 0.09 \\
ER2 & Ledger Satisfaction               & 0.25 \\
ER3 & Clip Grounding and Relevance      & 0.22 \\
ER4 & Timeline and Segment Coherence    & 0.11 \\
ER5 & Duration and Pacing Alignment     & 0.15 \\
ER6 & Music / Beat Alignment            & 0.05 \\
ER7 & Edit Economy / Minimality         & 0.11 \\
\bottomrule
\end{tabular*}
\caption{Average rubric weights across the train set (rounded;
per-task weights sum to $1$). Family A puts more mass on ER4;
families B and C put more mass on ER3.}
\label{tab:rubric-weights}
\end{table}

\paragraph{Candidate generation.}
States with sampled candidates: $1{,}500$. Total candidates:
$6{,}000$ ($K{=}4$ per state). JSON parse success $0.9968$;
patch-apply success $0.9960$.

\paragraph{Rubric scoring.}
Mean rubric score $0.67$ (std $0.13$). The candidate pool shows a
high verifier / rubric correlation $\rho = 0.94$, which indicates
that the rubric mostly tracks the verifier signal on these
states. The joint score used for pair construction is
$S(c)=0.65V(c)+0.35R(c)$, making the self-improvement stage
verifier-centered while retaining rubric-structured margins.
Criterion ER1 (intent) and the judge-side component of ER3
(clip relevance, judge-side) fall back to deterministic proxies
when external judge features are unavailable; the
\texttt{fallback\_used} flag is recorded for every one of the
$6{,}000$ candidates. In this run, both channels used the
constant neutral value $0.5$ on $6{,}000/6{,}000$ candidates.
Because the $K{=}4$ candidates of a state receive the same
constant, it cancels in every within-state score difference, so
the fallback shifts absolute rubric values without affecting
candidate selection, margins, or preference-pair construction; a
counterfactual rescoring with both channels zeroed reproduces
every ranking and every training pair.

\paragraph{Pair construction.}
$779$ chosen / rejected pairs are kept after applying a fixed
margin threshold. Mean margin $0.18$; median margin $0.19$;
hard-negative rate $0.987$. The training set is oversampled to
$1{,}651$ rows. The number of training states skipped because
the highest-scoring valid repair and the lower-scoring hard
negative fell inside the margin band is $635$.

\paragraph{Ablation caveats.}
Under the matched ablation (Table~\ref{tab:ablation}),
verifier-only DPO reaches $0.909$ VES against $0.924$ for the
full method: the main signal is verifier-scored student
self-improvement, while rubric-structured deterministic scoring
adds $0.015$ and improves margin construction and
interpretability. Temporal contrast was not trained: the data
summary records $0$ temporal-contrast pairs.

\section{Additional Results}
\label{app:additional}

\paragraph{Teacher replay diagnostics (RQ2).}

\begin{center}
\begin{minipage}{\columnwidth}
\centering
\scriptsize
\setlength{\tabcolsep}{2.2pt}
\renewcommand{\arraystretch}{1.05}
\begin{adjustbox}{max width=\columnwidth}
\begin{tabular}{@{}lrrrrrrrrr@{}}
\toprule
\textbf{Teacher} & \textbf{$N$} & \textbf{Branches} &
\textbf{FCSR} & \textbf{Hard} & \textbf{PASR} &
\textbf{ReqCR} & \textbf{NoReg} & \textbf{Dur} & \textbf{VES} \\
\midrule
GPT-5.4         & 2{,}000 & 6{,}000 & 0.379 & 0.015 & 0.872 & 0.698 & 1.000 & 0.202 & 0.521 \\
Qwen3-Max       & 1{,}946 & 3{,}892 & 0.446 & 0.016 & 0.989 & 0.712 & 1.000 & 0.157 & 0.557 \\
DeepSeek-V4-Pro & 1{,}959 & 3{,}920 & 0.473 & 0.028 & 0.946 & 0.703 & 1.000 & 0.156 & 0.559 \\
\bottomrule
\end{tabular}
\end{adjustbox}
\captionsetup{hypcap=false}
\captionof{table}{\textbf{Extended teacher replay diagnostics.}
Expands Table~\ref{tab:teacher-main} with NoRegression and
Dur@2s. This diagnoses the teacher trajectory resource used for
distillation (Section~\ref{sec:exp-rq2}), not student closed-loop
performance; same-loop teacher policies are in
Table~\ref{tab:teacher-loop}.}
\label{tab:teacher}
\end{minipage}
\end{center}

\paragraph{Cross-backbone transfer (RQ5a).}
The transfer runs repeat the Prompted\,/\,Raw\,/\,Verified
comparison on Llama-3.1-8B-Instruct and GLM-4-9B with the same
training data, hyperparameters, and frozen closed-loop protocol.
We fixed the success criterion (Verified$-$Raw $\ge +0.10$ VES
with same-direction HardPass) before running, and first checked
that each backbone emits parseable, applicable step-1 patches
($\ge 7/10$; Llama $8/10$, GLM $10/10$).
Table~\ref{tab:backbone-transfer} reports the outcome: the
Verified$-$Raw gain is significantly positive on all three
families and exceeds $+0.10$ on two; the GLM gain is smaller but
significant.
Unfiltered imitation is flat or harmful (Raw$-$Prompted $+0.026$,
$-0.070$, $-0.035$); on GLM the raw data also lowers patch-apply
success from $0.95$ to $0.75$. Required-clip recall rises from
$0.72/0.39/0.77$ to $0.98/0.92/0.99$.

\begin{center}
\begin{minipage}{\columnwidth}
\centering
\scriptsize
\renewcommand{\arraystretch}{1.05}
\begin{adjustbox}{max width=\columnwidth}
\begin{tabular}{@{}lrrrlr@{}}
\toprule
\textbf{Backbone} & \textbf{Prompt.} & \textbf{Raw} &
\textbf{Verif.} & \textbf{$\Delta$V$-$R [95\% CI]} & \textbf{$p$} \\
\midrule
Qwen3-8B     & 0.594 & 0.620 & \best{0.858} & $+0.238$ & -- \\
Llama-3.1-8B & 0.566 & 0.496 & \best{0.649} & $+0.153$ $[0.109,0.197]$ & $3.6\mathrm{e}{-8}$ \\
GLM-4-9B     & 0.643 & 0.607 & \best{0.686} & $+0.079$ $[0.036,0.120]$ & $9.0\mathrm{e}{-4}$ \\
\bottomrule
\end{tabular}
\end{adjustbox}
\captionsetup{hypcap=false}
\captionof{table}{\textbf{Cross-backbone transfer of
verifier-replayed SFT.} Same data, recipe, and frozen protocol;
paired statistics on Common-100. The Qwen3-8B row is the primary
progression of Table~\ref{tab:main}.}
\label{tab:backbone-transfer}
\end{minipage}
\end{center}

\paragraph{Same-loop teacher and frontier policies (RQ5b).}
Each teacher runs as an online policy in the identical closed
loop: the same task records, typed state, ledger, RefinePatch
schema, parser, Apply/Verify, feedback, $T{=}3$ budget, and
CSR$\ge$$0.8$ stopping rule, with no weight updates.
Table~\ref{tab:teacher-loop} reports final scores, latency, and
paired differences against RefineCut-Evo. Newer-generation
frontier policies under the same contract score $0.940$
(gpt-5.4-mini), $0.943$ (Qwen3.5-397B), $0.936$ (its
workflow-reflection variant), and $0.933$ (deepseek-v4-flash).
Existing workflow video-editing agents assume tool interfaces
different from our typed timeline, ledger, and RefinePatch loop
and cannot be run in it directly, so the workflow-reflection
variant stands in for that family here. Per-task call
records are released with the artifacts.

\begin{center}
\begin{minipage}{\columnwidth}
\centering
\scriptsize
\setlength{\tabcolsep}{2.4pt}
\renewcommand{\arraystretch}{1.05}
\begin{adjustbox}{max width=\columnwidth}
\begin{tabular}{@{}lrrrrlr@{}}
\toprule
\textbf{Policy} & \textbf{VES} & \textbf{Hard} & \textbf{Cvg} &
\textbf{s/task} & \textbf{Evo$-$policy [95\% CI]} & \textbf{W/T/L} \\
\midrule
GPT-5.4         & 0.893 & 0.76 & 0.86 & 6.7  & $+0.030$ $[0.001,0.062]$ & 16/74/10 \\
Qwen3-Max       & 0.773 & 0.63 & 0.72 & 32.8 & $+0.150$ $[0.091,0.213]$ & 30/61/9 \\
DeepSeek-V4-Pro & \best{0.936} & \best{0.89} & \best{0.95} & 23.2 & $-0.012$ $[-0.034,0.010]$ & 6/82/12 \\
\rowcolor{bandgreen}
\textbf{RefineCut-Evo (8B, local)} & 0.924 & 0.82 & \best{0.95} & 11.7 & -- & -- \\
\bottomrule
\end{tabular}
\end{adjustbox}
\captionsetup{hypcap=false}
\captionof{table}{\textbf{Same-loop teacher policies.} Identical
closed-loop contract; differences are mean paired per-task
values, and W/T/L counts are per-task Evo wins, ties, and losses.
The DeepSeek difference is not significant.}
\label{tab:teacher-loop}
\end{minipage}
\end{center}

\paragraph{Prompting, feedback, and search over the untrained
backbone (RQ5c).}
All conditions in Table~\ref{tab:ladder} run the untrained
Qwen3-8B backbone under the frozen protocol with the stopping
rule matched to the RefineCut-Evo loop. Verifier feedback adds $+0.090$ over direct
prompting, and $R{=}4$ reranking adds $+0.054$ over $R{=}1$
sampling. The visited-pool oracle selects, per task, the best
state visited by any of these runs; even this upper bound stays
$0.212$ VES below RefineCut-Evo (95\% CI $[0.170,0.252]$). The
feedback condition corresponds to the Prompted row of
Table~\ref{tab:main}; under the fully matched stopping rule,
tasks continue to the shared horizon instead of stopping at a
patch-extraction failure, giving $0.592$ against $0.594$, with
all state-level components identical.

\begin{center}
\begin{minipage}{\columnwidth}
\centering
\scriptsize
\renewcommand{\arraystretch}{1.05}
\begin{tabular*}{\columnwidth}{@{\extracolsep{\fill}}lrrr@{}}
\toprule
\textbf{Qwen3-8B condition} & \textbf{VES} & \textbf{Hard} & \textbf{Cvg} \\
\midrule
Direct prompting ($T{=}1$)        & 0.502 & 0.07 & 0.08 \\
$+$ verifier feedback ($T{\le}3$) & 0.592 & 0.15 & 0.21 \\
Sampling $R{=}1$                  & 0.646 & 0.23 & 0.30 \\
$R{=}4$ verifier reranking        & 0.700 & 0.32 & 0.42 \\
Visited-pool oracle               & 0.712 & 0.32 & 0.40 \\
\rowcolor{bandgreen} \textbf{RefineCut-Evo} & \best{0.924} & \best{0.82} & \best{0.95} \\
\bottomrule
\end{tabular*}
\captionsetup{hypcap=false}
\captionof{table}{\textbf{Prompting and search ladder over the
untrained backbone}, including the $R{=}1$ sampling row omitted
from Table~\ref{tab:sameloop}.}
\label{tab:ladder}
\end{minipage}
\end{center}

\paragraph{dev100 checkpoint selection.}

\begin{center}
\begin{minipage}{\columnwidth}
\centering
\scriptsize
\renewcommand{\arraystretch}{1.05}
\begin{adjustbox}{max width=\columnwidth}
\begin{tabular}{@{}lccccc@{}}
\toprule
\textbf{Model} & \textbf{FCSR} & \textbf{Hard} & \textbf{Dur} & \textbf{Cvg} & \textbf{VES} \\
\midrule
Verified               & 0.473 & 0.240 & 0.360 & 0.260 & 0.496 \\
Mixed-Pref             & 0.534 & 0.260 & 0.400 & 0.310 & 0.543 \\
\rowcolor{bandgreen} \textbf{Evo step300} & \best{0.608} & \best{0.450} & \best{0.570} & \best{0.470} & \best{0.604} \\
Evo step600            & 0.590 & 0.440 & 0.570 & 0.470 & 0.587 \\
\bottomrule
\end{tabular}
\end{adjustbox}
\captionsetup{hypcap=false}
\captionof{table}{\textbf{dev100 checkpoint selection for
RefineCut-Evo.} Step-$300$ is selected by dev VES;
step-$600$ shows mild over-optimization.}
\label{tab:dev-selection}
\end{minipage}
\end{center}

\paragraph{Paired significance.}

\begin{center}
\begin{minipage}{\columnwidth}
\centering
\scriptsize
\renewcommand{\arraystretch}{1.05}
\begin{adjustbox}{max width=\columnwidth}
\begin{tabular}{@{}lccccc@{}}
\toprule
\textbf{Subset} & \textbf{vs.} & \textbf{$\Delta$VES} & \textbf{95\% CI} & \textbf{W/T/L} & \textbf{$p$} \\
\midrule
\rowcolor{bandgreen} Common-100 & Mixed-Pref     & $+0.059$ & $[0.028, 0.092]$ & 22/73/5 & $0.0015$ \\
Common-100 & Verified       & $+0.066$ & $[0.036, 0.096]$ & 24/71/5 & $0.0005$ \\
Common-100 & Traj-Pref      & $+0.060$ & $[0.029, 0.091]$ & 22/73/5 & $0.0015$ \\
\midrule
\rowcolor{bandgreen} Canonical-clean & Mixed-Pref     & $+0.058$ & $[0.027, 0.090]$ & 21/66/5 & $0.0025$ \\
Canonical-clean & Verified       & $+0.068$ & $[0.036, 0.100]$ & 23/64/5 & $0.0009$ \\
Canonical-clean & Traj-Pref      & $+0.059$ & $[0.028, 0.091]$ & 21/66/5 & $0.0025$ \\
\bottomrule
\end{tabular}
\end{adjustbox}
\captionsetup{hypcap=false}
\captionof{table}{\textbf{Paired significance for RefineCut-Evo.}
$\Delta$VES is the mean paired per-task difference, not the
aggregate table-level difference; the sign-test $p$-value treats
ties as halves.}
\label{tab:paired}
\end{minipage}
\end{center}

\paragraph{Closed-loop failure taxonomy.}

\begin{center}
\begin{minipage}{\columnwidth}
\centering
\scriptsize
\renewcommand{\arraystretch}{1.05}
\begin{tabular*}{\columnwidth}{@{\extracolsep{\fill}}lccc@{}}
\toprule
\textbf{Outcome} & \textbf{Mixed-Pref} & \textbf{RefineCut-Evo} & \textbf{Change} \\
\midrule
\rowcolor{bandgreen} OK plan & 74 & \best{91} & \poschg{+17} \\
DurationMismatch & 16 & \best{1} & \poschg{-15} \\
LowCSRNoSingleFailure & 7 & \best{4} & \poschg{-3} \\
MissingClipReference & 3 & 3 & 0 \\
PatchApplyFail & 0 & 1 & \negchg{+1} \\
\bottomrule
\end{tabular*}
\captionsetup{hypcap=false}
\captionof{table}{\textbf{Closed-loop failure taxonomy on Common-100.}
The compact view highlights the main shift: RefineCut-Evo turns
more tasks into fully OK plans, primarily by removing duration
mismatches. Counts are task counts out of $100$.}
\label{tab:failure}
\end{minipage}
\end{center}

\paragraph{Evo-degraded tasks.}
Of the five tasks where RefineCut-Evo scores below Mixed-Pref
(Table~\ref{tab:paired}), four preserve execution, duration, and
convergence and each misses exactly one constraint entry: a
required closing clip, a semantic tag, a transition style, and a
required kept clip (three family-A composition tasks and one
family-B local repair). One family-C generative-assembly task
fails more broadly, with patch applicability dropping to $0.33$
and a duration overshoot. The degradations are local
single-constraint misses rather than a systematic capability
loss.

\paragraph{Canonical-clean closed-loop.}

\begin{center}
\begin{minipage}{\columnwidth}
\centering
\scriptsize
\renewcommand{\arraystretch}{1.05}
\begin{tabular*}{\columnwidth}{@{\extracolsep{\fill}}lcccccc@{}}
\toprule
\textbf{Model} & \textbf{N} & \textbf{FCSR} & \textbf{Hard} & \textbf{Dur} & \textbf{Cvg} & \textbf{VES} \\
\midrule
Prompted                            & 92 & 0.607 & 0.152 & 0.652 & 0.217 & 0.591 \\
Raw                                 & 92 & 0.552 & 0.152 & 0.424 & 0.228 & 0.612 \\
Verified                            & 92 & 0.872 & 0.609 & 0.804 & 0.772 & 0.849 \\
Traj-Pref                           & 92 & 0.872 & 0.652 & 0.837 & 0.783 & 0.858 \\
Mixed-Pref                          & 92 & 0.879 & 0.652 & 0.826 & 0.793 & 0.859 \\
\rowcolor{bandgreen} \textbf{RefineCut-Evo} & 92 & \best{0.949} & \best{0.804} & \best{0.978} & \best{0.946} & \best{0.917} \\
\bottomrule
\end{tabular*}
\captionsetup{hypcap=false}
\captionof{table}{Closed-loop results on the canonical-clean subset of
Common-100 ($N=92$).}
\label{tab:canon-clean-full}
\end{minipage}
\end{center}

\paragraph{Per-family breakdown.}

\begin{center}
\begin{minipage}{\columnwidth}
\centering
\scriptsize
\renewcommand{\arraystretch}{1.05}
\begin{tabular*}{\columnwidth}{@{\extracolsep{\fill}}lcc@{}}
\toprule
\textbf{Family} & \textbf{Mixed-Pref VES} & \textbf{RefineCut-Evo VES} \\
\midrule
A (composition)                    & 0.865 & \best{0.892} \\
B (insert / repair)                & 0.871 & \best{0.975} \\
C (generative)                     & 0.853 & \best{0.890} \\
\bottomrule
\end{tabular*}
\captionsetup{hypcap=false}
\captionof{table}{Per-family closed-loop VES on Common-100. Per-subtype
counts are released in the supplementary artifacts.}
\label{tab:by-family}
\end{minipage}
\end{center}

\paragraph{Plan statistics and reward-hacking diagnostics.}

\begin{center}
\begin{minipage}{\columnwidth}
\centering
\scriptsize
\renewcommand{\arraystretch}{1.05}
\begin{tabular*}{\columnwidth}{@{\extracolsep{\fill}}lcccc@{}}
\toprule
\textbf{Model} & \makecell{\textbf{Patches}\\\textbf{/task}} &
\makecell{\textbf{Ops}\\\textbf{/task}} &
\makecell{\textbf{Op.}\\\textbf{entropy}} &
\makecell{\textbf{Output}\\\textbf{length}} \\
\midrule
Prompted    & 2.72 & 5.95 & 0.83 & 69.1 \\
Verified    & 2.61 & 6.23 & 0.89 & 131.5 \\
Mixed-Pref  & 2.58 & 6.11 & 0.89 & 135.0 \\
\rowcolor{bandgreen} \textbf{RefineCut-Evo} & \best{1.76} & \best{3.77} & \best{0.99} & 67.5 \\
\bottomrule
\end{tabular*}
\captionsetup{hypcap=false}
\captionof{table}{Per-variant plan statistics on Common-100.
Invalid clip references and over-rewrites are $0$ for all listed
variants, no-regression stays $1.0$, and RefineCut-Evo's mean
duration deviation drops from $1.55$\,s (Mixed-Pref) to
$0.34$\,s. RefineCut-Evo improves with fewer, more selective
edits, not longer outputs.}
\label{tab:planstats}
\end{minipage}
\end{center}

\paragraph{Semantic-input intervention (full deltas).}
An unmodified control run reproduces the headline VES ($0.9237$). Removing only
the short caption, while retaining the structured
subject/action/scene/motion fields, yields $0.9190$ (paired
clean-minus-condition $+0.0048$, 95\% CI $[-0.0145,0.0255]$,
n.s.). Removing all clip semantics yields $0.8502$ ($+0.0736$,
$[0.0422,0.1093]$, $p=4.9\mathrm{e}{-5}$), and shuffling semantic
descriptions across clips yields $0.7241$ ($+0.1996$,
$[0.1579,0.2431]$, $p=1.6\mathrm{e}{-15}$), with required-clip
recall falling from $0.98$ to $0.46$.

\paragraph{Distillation-stage API-judge analysis.}
\label{app:judge-validity}

Three independent blind API judges are run on distillation-stage
variants using a stripped pairwise prompt that contains no model
identifiers, file paths, or judge-identifying text (full prompt
in Appendix~\ref{app:prompts}). Per-judge raw counts on
the $20$-pair Common-100 panel are in Table~\ref{tab:judge-wtl};
final-stage alignment with the verifier is analyzed below. Pooled
non-tie win rates, aggregated across judges, are $0.962$ for
Verified~vs.\ Raw, $0.880$ for Mixed-Pref~vs.\ Verified, $0.764$
for Mixed-Pref~vs.\ Prompted, $0.805$ for Traj-Pref~vs.\ Verified,
and $0.655$ for Traj-Pref~vs.\ Prompted.

\begin{center}
\begin{minipage}{\columnwidth}
\centering
\scriptsize
\renewcommand{\arraystretch}{1.05}
\begin{tabular*}{\columnwidth}{@{\extracolsep{\fill}}l c ccc c c@{}}
\toprule
\textbf{Judge} & \textbf{Comparison} & \textbf{W} & \textbf{T} & \textbf{L} & \textbf{$n$} & \textbf{WR$_\text{ex}$} \\
\midrule
Judge A   & Verified vs.\ Raw       & 17 & 2  & 1 & 20 & 0.944 \\
Judge B$^{\dagger}$ & Verified vs.\ Raw & 18 & 1 & 0 & \textbf{19} & 1.000 \\
Judge C      & Verified vs.\ Raw       & 16 & 3  & 1 & 20 & 0.941 \\
\midrule
Judge A   & Mixed vs.\ Verified     & 8  & 10 & 2 & 20 & 0.800 \\
Judge B$^{\dagger}$ & Mixed vs.\ Verified & 6 & 13 & 0 & \textbf{19} & 1.000 \\
Judge C      & Mixed vs.\ Verified     & 8  & 11 & 1 & 20 & 0.889 \\
\midrule
Judge A   & Mixed vs.\ Prompted         & 15 & 1  & 4 & 20 & 0.789 \\
Judge B   & Mixed vs.\ Prompted         & 14 & 2  & 4 & 20 & 0.778 \\
Judge C      & Mixed vs.\ Prompted         & 13 & 2  & 5 & 20 & 0.722 \\
\midrule
Judge A   & Traj-Pref vs.\ Verified & 11 & 7  & 2 & 20 & 0.846 \\
Judge B   & Traj-Pref vs.\ Verified & 10 & 6  & 4 & 20 & 0.714 \\
Judge C      & Traj-Pref vs.\ Verified & 12 & 6  & 2 & 20 & 0.857 \\
\midrule
Judge A   & Traj-Pref vs.\ Prompted     & 10 & 2  & 8 & 20 & 0.556 \\
Judge B   & Traj-Pref vs.\ Prompted     & 14 & 0  & 6 & 20 & 0.700 \\
Judge C      & Traj-Pref vs.\ Prompted     & 14 & 0  & 6 & 20 & 0.700 \\
\bottomrule
\end{tabular*}
\captionsetup{hypcap=false}
\captionof{table}{Per-(judge, comparison) raw counts for the
distillation-stage blind API-judge analysis.
$\mathrm{WR}_\text{ex} = W/(W+L)$. $^{\dagger}$Judge B parse
coverage is $n{=}19$ on these two rows.}
\label{tab:judge-wtl}
\end{minipage}
\end{center}

\paragraph{Final-stage Evo API-judge alignment.}
A blind API judge was run on final-stage comparisons.
After parse filtering, the aggregate non-tie win rates are
$0.574$ for Evo vs.\ Mixed-Pref ($108$W/$108$T/$80$L), $0.545$
for Evo vs.\ Verified ($103$W/$110$T/$86$L), and $0.548$ for
Evo vs.\ Traj-Pref ($102$W/$108$T/$84$L). Aligning per-task
judgments with the verifier margin explains this structure:
judge ties concentrate on tasks with $\Delta\mathrm{VES}=0$
(mean $|\Delta\mathrm{VES}|\approx 0.00$ on judge-tie tasks vs.\
${\approx}0.13$ on non-tie tasks; Mann--Whitney
$p\le 1.3\mathrm{e}{-5}$, consistent across the three
comparisons); when both sides give a non-tie verdict, directions
agree on $68$--$70\%$ of tasks, and on every task the verifier
scores as an Evo degradation the judge points the same way. On
$\Delta\mathrm{VES}=0$ tasks the judge's non-tie verdicts are
near-symmetric (e.g., $16$W/$20$L). The high tie rate thus
reflects genuinely equivalent terminal plans ($73/100$ tasks at
$\Delta\mathrm{VES}=0$) rather than disagreement between the
evaluation layers; rendered previews amplify perceptual
differences and separate the same comparisons more strongly
(Appendix~\ref{app:human}).

\paragraph{Step-Pref.}
A step-level-only preference variant (Step-Pref) is
inconclusive: gains over Verified on the closed-loop
summary are within the variance of the multi-seed runs we
performed.

\FloatBarrier
\section{Human Evaluation and Human-Written Brief Validation}
\label{app:human}

This appendix reports the blind rendered-preview evaluation
(Section~\ref{sec:exp-human}) and the Human50 human-written brief
validation. Human evaluators compare blind
randomized A/B rendered storyboard previews. Each comparison has
$150$ judged pairs and three annotators per pair. Preference score is
$(W+0.5T)/N$ for the first model in each comparison. Annotators
were recruited from university campuses and partner companies,
all with one to five years of video-editing experience, and were
compensated at a rate consistent with local standards for their
region. They received a written instruction sheet covering the
rating axes, blinded protocol, compensation, opt-out policy, and
a brief content-risk disclaimer; the full instruction text is
released with the supplementary artifacts.

\begin{center}
\begin{minipage}{\columnwidth}
\centering
\scriptsize
\setlength{\tabcolsep}{2.4pt}
\renewcommand{\arraystretch}{1.05}
\begin{tabular*}{\columnwidth}{@{\extracolsep{\fill}}lcccccc@{}}
\toprule
\textbf{Comparison} & \textbf{W/T/L} &
\makecell{\textbf{Pref.}\\\textbf{Score}} &
\makecell{\textbf{Non-tie}\\\textbf{WR}} &
\makecell{\textbf{Maj.}\\\textbf{Agree.}} &
\textbf{$\kappa$} & \textbf{$\alpha$} \\
\midrule
\rowcolor{bandgreen}
Evo vs Mixed & 100/34/16 & \best{0.780} & \best{0.862} & 0.873 & 0.620 & 0.580 \\
Evo vs Traj  & 95/37/18  & 0.757 & 0.841 & 0.860 & 0.590 & 0.553 \\
Evo vs Verified & 97/35/18  & 0.763 & 0.843 & 0.867 & 0.601 & 0.562 \\
Mixed vs Prompted & 124/18/8 & 0.887 & 0.939 & 0.927 & 0.781 & 0.742 \\
\bottomrule
\end{tabular*}
\captionsetup{hypcap=false}
\captionof{table}{\textbf{Final pairwise human evaluation}
($N=150$ per comparison, three annotators per pair). Higher
preference scores and non-tie win rates prefer the first model.
The sanity comparison confirms that the protocol separates a known
large quality gap.}
\label{tab:human-pairwise-full}
\end{minipage}
\end{center}

\paragraph{Human50 construction.}
Human50 contains $50$ free-form briefs written independently by
seven contributors unaffiliated with the authors (three vloggers,
four editing students) under a shared instruction sheet. For each
brief, an LLM proposed candidate ledger entries; two research
assistants revised them into decidable constraint entries
following the public construction guide, and two independent
professional video editors validated every entry. The briefs are
free-form, while the ledgers retain the executable constraint
vocabulary of the frozen verifier, so the instruction source is
human and evaluation stays decidable.
Table~\ref{tab:human-curated} reports closed-loop results on this
set.

\begin{center}
\begin{minipage}{\columnwidth}
\centering
\scriptsize
\setlength{\tabcolsep}{3.0pt}
\renewcommand{\arraystretch}{1.05}
\begin{tabular*}{\columnwidth}{@{\extracolsep{\fill}}lcccccc@{}}
\toprule
\textbf{Model} & \textbf{FCSR} & \textbf{Hard} &
\textbf{Dur} & \textbf{Cvg} & \textbf{ReqCR} &
\textbf{VES} \\
\midrule
Prompted & 0.582 & 0.140 & 0.620 & 0.180 & 0.270 & 0.578 \\
Raw & 0.534 & 0.160 & 0.420 & 0.220 & 0.710 & 0.608 \\
Verified & 0.856 & 0.580 & 0.780 & 0.740 & 0.960 & 0.839 \\
Traj-Pref & 0.852 & 0.620 & 0.800 & 0.740 & 0.960 & 0.842 \\
Mixed-Pref & 0.864 & 0.620 & 0.800 & 0.760 & 0.960 & 0.848 \\
\rowcolor{bandgreen}
RefineCut-Evo & \best{0.928} & \best{0.760} & \best{0.940} & \best{0.900} & \best{0.980} & \best{0.902} \\
\bottomrule
\end{tabular*}
\captionsetup{hypcap=false}
\captionof{table}{\textbf{Human-written brief validation}
($N=50$). The ranking from Common-100 is preserved; the Evo-stage
gap remains $+0.054$ VES over Mixed-Pref.}
\label{tab:human-curated}
\end{minipage}
\end{center}

\FloatBarrier

\section{Full Prompts}
\label{app:prompts}

This appendix collects compact versions of the six prompts used
throughout the pipeline. The full text of each prompt is included
in the supplementary artifacts; each entry below records
the prompt role, expected inputs, output schema, and a compact
excerpt for readability.

\paragraph{Task-generation prompt.}
\textbf{Purpose:} generate benchmark tasks and ledgers from clip
pools. \textbf{Input fields:} clip pool, task family, subtype,
ledger template. \textbf{Output schema:} task record JSON.
\textbf{Version/source path:} supplementary prompt artifact;
compact excerpt below.
\begin{promptbox}[pborange]{Task-generation prompt (compact)}
system: You are a video-editing task author.
task: Given a clip pool and a family/subtype slot,
emit one task record with fields
  task_id, task_type, task_subtype,
  brief, target_duration,
  constraint_ledger (>= 4 items, includes target_duration),
  clip_pool, structure, subtype_constraints.
constraints:
  - clip_ids in clip_pool MUST come from given_pool.
  - clip_pool size in [30, 60] when given_pool>=30.
  - ledger items use the 14 allowed constraint types.
output: STRICT JSON; no commentary.
\end{promptbox}

\paragraph{Caption prompt.}
\textbf{Purpose:} convert sampled clip frames into metadata read
by the planner. \textbf{Input fields:} clip frames, motion bucket,
clip id, source. \textbf{Output schema:} one caption metadata row.
\textbf{Version/source path:} supplementary prompt artifact;
compact excerpt below.
\begin{promptbox}[pbgreen]{Caption prompt (compact)}
system: You are a video-clip captioner.
input: a clip (sampled frames; sampling rate
  determined by motion bucket: low=3, med=5, high=7).
task: emit one JSON row with fields
  clip_id, source, duration, subject, action,
  scene, camera, scene_category,
  motion_intensity, caption_short,
  model_used (record which captioner produced this row),
  ts (ISO timestamp).
output: STRICT JSON; no commentary.
\end{promptbox}

\paragraph{Trajectory-teacher prompt.}
\textbf{Purpose:} collect frontier-teacher candidate repair
branches. \textbf{Input fields:} brief, clip pool, music metadata,
timeline state, constraint ledger. \textbf{Output schema:} four
RefinePatch candidates. \textbf{Version/source path:}
supplementary prompt artifact; compact excerpt below.
\begin{promptbox}[pbpink]{Trajectory-teacher prompt (compact)}
system: You are a frontier video-editing planner.
input: brief, clip_pool with captions and metadata,
  music metadata, current timeline state,
  constraint_ledger (with per-item satisfied flags).
task: at this refine step, emit EXACTLY 4 candidate
  RefinePatch branches. Each branch must contain
  operations[], rationale_against_ledger[],
  repair_operator, thought, confidence.
output: a list of 4 RefinePatch objects in STRICT JSON.
\end{promptbox}

\paragraph{Unified PatchPlanner prompt.}
\textbf{Purpose:} run the trained planner in closed-loop
evaluation. \textbf{Input fields:} brief, clip pool, captions,
music metadata, current state, ledger, violated subset.
\textbf{Output schema:} one RefinePatch. \textbf{Version/source
path:} supplementary prompt artifact; compact excerpt below.
\begin{promptbox}[pbblue]{Unified PatchPlanner prompt (compact, used at test)}
system: You are an executable video-editing planner.
input: brief, clip_pool, captions, music_metadata,
  current_state, constraint_ledger (with satisfied flags
  and a violated_subset listing currently unmet items).
task: emit one RefinePatch that targets ledger items
  in violated_subset and does not regress satisfied items.
constraints:
  - operations limited to RFC 6902 add/remove/replace
    on canonical paths of current_state.
  - reference only clip_ids that appear in clip_pool.
  - keep rationale_against_ledger aligned with operations.
output: STRICT JSON; no commentary.
\end{promptbox}

\paragraph{Rubric construction template.}
\textbf{Purpose:} build task-specific ER1--ER7 scoring weights for
RefineCut-Evo. \textbf{Input fields:} task family and subtype.
\textbf{Output schema:} rubric JSON with criteria, weights, and
fallbacks. \textbf{Version/source path:} supplementary prompt
artifact; compact excerpt below.
\begin{promptbox}[pbpurple]{Rubric construction template (compact)}
template: For task_subtype X under family F:
  - assign ER1..ER7 a non-negative weight; sum to 1.
  - default weights start from the family template
    (A puts mass on ER4; B and C put mass on ER3).
  - for each criterion list:
      criterion_name, weight,
      verifier_signals_used,
      fallback_proxy_if_judge_absent.
output: a rubric JSON used by the Evo scorer.
\end{promptbox}

\paragraph{Stripped blind judge prompt.}
\textbf{Purpose:} audit distillation-stage plans without model
identity leakage. \textbf{Input fields:} brief, ledger, plan A,
plan B. \textbf{Output schema:} strict JSON winner plus reason.
\textbf{Version/source path:} supplementary prompt artifact;
compact excerpt below.
\begin{promptbox}[pbteal]{Stripped blind judge prompt (compact)}
system: You are a blind video-editing-plan judge.
You will see two plans for the SAME task.
The model identity and any file path are removed.
input: brief, ledger, plan_A, plan_B.
task: pick the plan that better satisfies the brief and
  the ledger; output one of {A, B, tie} with a brief
  reason. Do not reveal the model identity in your
  reasoning. Do not mention "as an LLM".
output: STRICT JSON {"winner": "A"|"B"|"tie",
                     "reason": "..."}.
\end{promptbox}

\section{Verifier Implementation}
\label{app:verifier}

The verifier is deterministic: it validates a patch, applies it
to the current state, recomputes the ledger, and emits both gate
signals and branch-score components.

\begin{algorithm}[!ht]
\small
\caption{Verifier replay for one candidate patch.}
\label{alg:verifier}
\begin{algorithmic}[1]
\Require state $s_t$, patch $p_t$, task record $\tau$
\If{$p_t$ fails schema validation} \State \Return reject(schema) \EndIf
\State $s_{t+1}\gets\mathrm{Apply}(s_t,p_t)$
\If{$s_{t+1}$ is invalid} \State \Return reject(apply) \EndIf
\If{timeline is not renderable} \State \Return reject(timeline) \EndIf
\State $L_t\gets\mathrm{RecomputeLedger}(s_t,\tau)$
\State $L_{t+1}\gets\mathrm{RecomputeLedger}(s_{t+1},\tau)$
\State compute $\Delta$CSR, TargetedRepair, ReqClipRecall,
PASR, NoRegression, and Locality
\State score $\gets w_1\Delta$CSR $+w_2$TargetedRepair
$+w_3$ReqClipRecall $+w_4$PASR $+w_5$NoRegression
$+w_6$Locality
\State \Return accepted patch, new state, new ledger, score terms
\end{algorithmic}
\end{algorithm}

\vspace{0.4em}
\begin{center}
\begin{minipage}{\columnwidth}
\centering
\scriptsize
\setlength{\tabcolsep}{2.4pt}
\renewcommand{\arraystretch}{0.98}
\begin{tabularx}{\columnwidth}{@{}lX@{}}
\toprule
\textbf{Metric} & \textbf{Rule} \\
\midrule
ReqClipRecall denominator &
Number of required keep-clip items; tasks with none are excluded
from this component average. \\
NoRegression &
Items satisfied at $L_t$ must remain satisfied after the patch;
the all-steps variant conjoins this across the loop. \\
Converged@3 &
Fires when terminal constraint satisfaction reaches at least
$0.8$ within three repair steps, including early termination. \\
TargetedRepair &
Counts only listed ledger items that were unsatisfied before the
patch and become satisfied after replay. \\
VES &
Uses the fixed weights in Appendix~\ref{app:metrics}; undefined
ReqClipRecall tasks use the normalized remaining components. \\
\bottomrule
\end{tabularx}
\captionsetup{hypcap=false}
\captionof{table}{\textbf{Verifier denominator and aggregation rules.}
These rules keep branch scoring deterministic and avoid treating
undefined clip-recall denominators as successes or failures.}
\label{tab:verifier-rules}
\end{minipage}
\end{center}

\section{Worked Example}
\label{app:worked-example}

This compact example illustrates one verifier-replay step for a
\texttt{b\_roll\_insert} task. It is schematic and is not a
released-record identifier.

\begin{center}
\begin{minipage}{\columnwidth}
\centering
\scriptsize
\setlength{\tabcolsep}{2.4pt}
\renewcommand{\arraystretch}{1.05}
\begin{tabularx}{\columnwidth}{@{}lX@{}}
\toprule
\textbf{Item} & \textbf{Illustrative value} \\
\midrule
Brief & Insert a short b-roll cutaway between a speaker close-up
and a wide street shot. \\
Pool & C001 speaker close-up; C002 wide street at dusk; C007
traffic-light cutaway; C012 empty alley. \\
Initial state & sequence [C001, C002], durations [5.4, 10.0],
total 15.4s. The 13s duration target is violated; required clips
C001 and C002 are already present. \\
Branch 1 & Replace the sequence with [C001, C007, C002] and set
durations to [5.0, 3.0, 5.0]; operator \texttt{reselect\_clip}. \\
Branch 2 & Keep [C001, C002] but set durations to [4.0, 9.0];
operator \texttt{adjust\_parameter}. \\
\bottomrule
\end{tabularx}
\end{minipage}
\end{center}

\paragraph{Replay outcome.}
Branch~1 is selected: it fixes duration, preserves required
clips, inserts C007, and touches only ledger-relevant fields.
Branch~2 fixes duration but misses the b-roll intent, so it is
the hard negative. After Branch~1, all four ledger items pass and
the loop ends at step~$1$.

\end{document}